\documentclass[sigconf]{aamas}

\usepackage{graphicx}
\usepackage{algorithm}
\usepackage{algorithmic}
\usepackage{subcaption}
\usepackage{makecell}
\usepackage{tabularx}

\newtheorem{theorem}{Theorem}
\newtheorem{definition}{Definition}

\usepackage{balance} 

\acmSubmissionID{869}

\title[AAMAS-2025 Formatting Instructions]{Agential AI for Integrated Continual Learning, Deliberative Behavior, and Comprehensible Models}

\author{Zeki Doruk Erden}
\affiliation{
  \institution{École Polytechnique Fédérale de Lausanne}
  \city{Lausanne}
  \country{Switzerland}}
\email{zeki.erden@epfl.ch}

\author{Boi Faltings}
\affiliation{
  \institution{École Polytechnique Fédérale de Lausanne}
  \city{Lausanne}
  \country{Switzerland}}
\email{boi.faltings@epfl.ch}

\begin{abstract}
Contemporary machine learning paradigm excels in statistical data analysis, solving problems that classical AI couldn't. However, it faces key limitations, such as a lack of integration with planning, incomprehensible internal structure, and inability to learn continually. We present the initial design for an AI system, Agential AI (AAI), in principle operating independently or on top of statistical methods, designed to overcome these issues. AAI's core is a learning method that models temporal dynamics with guarantees of completeness, minimality, and continual learning, using component-level variation and selection to learn the structure of the environment. It integrates this with a behavior algorithm that plans on a learned model and encapsulates high-level behavior patterns. Preliminary experiments on a simple environment show AAI's effectiveness and potential.
\end{abstract}


\newcommand{\BibTeX}{\rm B\kern-.05em{\sc i\kern-.025em b}\kern-.08em\TeX}

\begin{document}


\pagestyle{fancy}
\fancyhead{}


\maketitle 


\section{Introduction}
\label{sec:intro}

The current machine learning (ML) paradigm uses continuous representations to approximate environmental structures through fixed internal architectures like neural networks (NNs). This approach has effectively addressed numerous challenges once considered among the toughest in AI, including vision \cite{khan2021machine}, language processing \cite{zhao2023survey}, and complex behavior \cite{li2017deep}. However, as these problems are solved, important limitations related to the methods of solving them and their practical integration into larger systems start to receive more attention \cite{clune2019ai,zador2019critique,marcus2018deep,lecun2022path}. In particular; these models, heavily overparameterized with finite expressive potential, adapt by tuning continuous parameters rather than learning the structure topologically. Consequently, information is embedded in a distributed manner, leading to several important issues that are widely regarded as core limitations of machine learning (and NNs, its current dominant paradigm) - most notably the incapability of continual learning and information reuse, incomprehensibility and non-designability of the internal structure, and difficulty integrating learned information with deliberative behavior.

These issues originate from the shared limitation of approximating environmental structures with fixed models, rather than learning them topologically. They can be addressed collectively and without limitations of individual subfields tackling them separately, through a different design philosophy that tackles the problem from the ground up. To that end, we present the initial design of a system called Agential AI (AAI). The system consists of three components: 

\begin{itemize}
    \item \textit{Modelleyen}, an alternative learning mechanism exemplifying what we call a \textit{varsel mechanism,} that captures the structure of the environment topologically in a discrete network without using gradients, enabling continual learning without destructive adaptation, and without task boundaries or replay,\footnote{An earlier version of Modelleyen has been presented in \cite{Erden2024Modelleyen}.}
    \item \textit{Planlayan}, a planning algorithm that executes goal-directed actions based on a model generated by Modelleyen,
    \item A \textit{behavior encapsulation mechanism}, currently demonstrated independently of agent operation, that decomposes behavior patterns produced by Planlayan into arbitrary hierarchical structures with autonomously detected subgoals.
\end{itemize}

We detail these components, explain how they overcome multiple major limitations of contemporary ML (detailed in the next section), and demonstrate their proof-of-principle operation on a simple test environment.

\section{Related Work}

\subsection{Common Limitations of ML Systems}

Two most important core limitations of current ML systems are the inability of continual learning and incomprehensibility of internal structure; often tackled in isolation \cite{kirkpatrick2017overcoming,rusu2016progressive,jacobson2022task,hadsell2020embracing,zhuang2020comprehensive,xu2019explainable}. These methods don't fully resolve the fundamental limitations of NNs but aim to mitigate their effects. For example, many continual learning solutions rely on assumptions that simplify the problem (e.g. externally defined task boundaries \cite{rusu2016progressive,jacobson2022task} or storage and replay of past observations \cite{buzzega2020dark}) or only bias learning towards past tasks without ensuring true continual learning \cite{kirkpatrick2017overcoming}. Similarly, Explainable AI methods \cite{xu2019explainable} attempt to provide post-hoc explanations for the operation of neural networks, yet they fail to address the fundamental incomprehensibility of their internal structures, leaving them far from being truly engineerable. Furthermore, the challenge of explainability is often approached independently of the continual learning problem, rather than being seen as stemming from a shared underlying issue. Even studies that consider explainability within the context of continual learning tend to either propose distinct, sequentially applicable mechanisms for each problem separately \cite{roy2022interpretable} or focus on additional explainability challenges that arise when certain continual learning methods are used \cite{rymarczyk2023icicle, cossu2024drifting}.

\subsection{Deliberative Behavior}

Planning is a well-established area of AI research \cite{ghallab2016automated}, offering advantages over reward-based learning for reactive behavior \cite{ccalicsir2019model}, as it is more precise and doesn’t require relearning for new goals. Traditional planning methods generally do not include environment model learning, and even those that do learn to model the environment to some degree face significant limitations. For instance, \cite{mordoch2023learning} operate under restrictive assumptions: \begin{quote}
(1) The conditions over the numeric state variables in actions’ preconditions are linear inequalities, (2) The numeric expressions defining actions’ effects are linear combinations of state variables, and (3) The set of numeric state variables involved in each action’s preconditions and effects are known in advance.
\end{quote} Analogous limitations can be found in other works as well \cite{verma2021asking, stern2017efficient}. Our modeling approach is not constrained by such assumptions.

While model-based reinforcement learning \cite{moerland2023model,moerland2020framework} partially addresses deliberative behavior through experience-driven learning, it suffers from limitations due to its non-structured representation of environments. This makes it challenging to represent alternative pathways to goals and conduct goal-oriented backward searches, often relying on random state sampling \cite{hammersley2013monte}. Our method’s planner explicitly represents alternative pathways using a learned model, enabling precise goal-directed behavior without the need for next-state sampling. 

\subsection{Behavior Decomposition}

A longstanding objective within the learning agents community has been to automatically break down behavior into distinct subunits, which is the primary motivation behind the subfield of Hierarchical Reinforcement Learning (HRL) \cite{pateria2021hierarchical}. However, this goal has yet to be achieved: current HRL methods produce rigid hierarchies that require predefining the structure in some form, with no exceptions known to us. Additionally, there is no existing capability for HRL-learned policies to be divided into multiple subpolicies, which is a fundamental requirement for flexible hierarchical structures. In this work, we present an initial demonstration of a behavior encapsulation mechanism (currently independent of the agent's operation) that can generate arbitrary hierarchical decompositions of behaviors designed by the planner. This mechanism can identify relevant subpolicies, along with their internal preconditions and subgoals, without any prior definitions, thus achieving the goal of HRL in a different context.

\begin{table*}[]
    \centering
    \caption{Main aims of current learning agents research, representative subfields tackling these aims, and inherent limitations of their approaches.}
    \begin{tabularx}{\linewidth}{X|X|X|X|X}
        \textbf{Aim} & Continual learning & Deliberative behavior & Behavior decomposition & Understandability-controllability \\
        \hline
        \textbf{Subfield} & Various & Model based RL & Hierarchical RL & Explainable AI \\
        \hline
        \textbf{Limits} & Requires either task boundaries or reexposure to past samples & Imprecise deliberation based on future-state sampling & Rigid prespecified hierarchy, subpolicies are not decomposable &  Post-hoc, keeps incomprehensible internal structure \\        
    \end{tabularx}
    \label{tab:aims_overview}
\end{table*}

\subsection{Summary}

Table \ref{tab:aims_overview} summarizes the previous discussion. As mentioned earlier, these issues arise from the shared limitation of approximating environmental structures with fixed models, rather than learning them structurally. Therefore, once this fundamental challenge is addressed, the issues can be tackled collectively. This is the central goal of this work. \footnote{Our approach to modelling is also possibly applicable to Bayesian structure learning \cite{kitson2023survey}; although this is not our primary motivation.}

\section{Modelleyen}
\label{sec:Modelleyen}

Modelleyen is designed to model sequential observations from an environment, but can be applied to any prediction task. It learns the environment’s structure with minimal exposure, enabling information reuse and continual learning while maintaining consistency with past experiences. At the core of our method is a local variation and selection process - an important fundamental property of biological systems that has not found their way explicitly into AI methods, whose importance in the generation of biological structures and facilitation of their further evolution \cite{gerhart2007theory,marc2005plausibility,west2003developmental}, including in the brain \cite{marc2005plausibility,edelman1993neural} has recently been particularly appreciated. As it will be clear, this mechanism essential to the realization of continual learning and structured environment modelling, which in turn leads to all the other capabilities.

Below, we outline Modelleyen’s core mechanism. We note in advance that the current version operates within a discrete state space and only accounts for immediate event succession without long-term relationship modeling (see Section \ref{sec:conclusion} for a discussion of these points). Due to space limitations, we provide only an overview of the key definitions, basic learning mechanism, and core continual learning properties. For a full description, the reader is referred to Appendix \ref{sec:modelleyen_details}, and Algorithms \ref{alg:algorithm_adaptationloop} and \ref{alg:algorithm_csvstate}.

\begin{definition}
    (State Variable - SV) A state variable $X$ is a unit in our system whose state, $S_X$, can take values 1 (\textit{active}), -1 (\textit{inactive}), or 0 (\textit{unobserved/undefined} depending on context). 
\end{definition}

SVs can be interpreted as boolean variables with additional possibility to take an additional "unobserved" value. The integers assigned for states are only for notation and not for algebraic operation. The following are subtypes of SVs:

\begin{definition}
    (Base SV - BSV and Dynamics SV - DSV) A \textit{BSV} $X$ is an SV whose values are provided externally each timestep and whose state is limited by $S_X\in\{-1,1\}$. Each BSV comes with two \textit{DSVs}, $X_A$ and $X_D$, that represent its activation and deactivation at current step ($t$) compared to previous timestep respectively; where $S_{X_A}=1$ if and only if $S_X(t-1)=-1 \land S_X(t)=1$, and $S_{X_D}=1$ if and only if $S_X(t-1)=1 \land S_X(t)=-1$, and persisting as long as no new event in BSVs are observed.
\end{definition}

\begin{definition}
    (Conditioning SV - CSV) A CSV $C$ is a type of SV with mutable sets of positive sources $X_P$, negative sources $X_N$, and conditioning targets $Y$. Positive and negative sources are BSVs and DSVs, while targets can be DSVs or other CSVs. The sources of $C$ are considered "satisfied" if all positive sources are active and all negative sources are not active. If sources are satisfied, $S_C=1$ if sources are satisfied and $S_Y \in \{0,1\},\ \forall x \in Y$ (targets are active); $S_C=-1$ if sources are satisfied and $S_Y \in \{0,-1\},\ \forall x \in Y$ (targets are inactive), and $S_C=0$ otherwise. Additionally, each CSV has a "unconditionality" flag, which indicates if the CSV has, in the past, been always observed active when sources were satisfied ("unconditional"), was never observed active without a predictive explanation ("conditional"), or was sometimes observed active without a predictive explanation ("possibly conditional"), the latter representing uncertainty in a qualitative manner.
    
\end{definition}

\begin{figure}[t]{}
     \centering
     \includegraphics[width=0.2\textwidth]{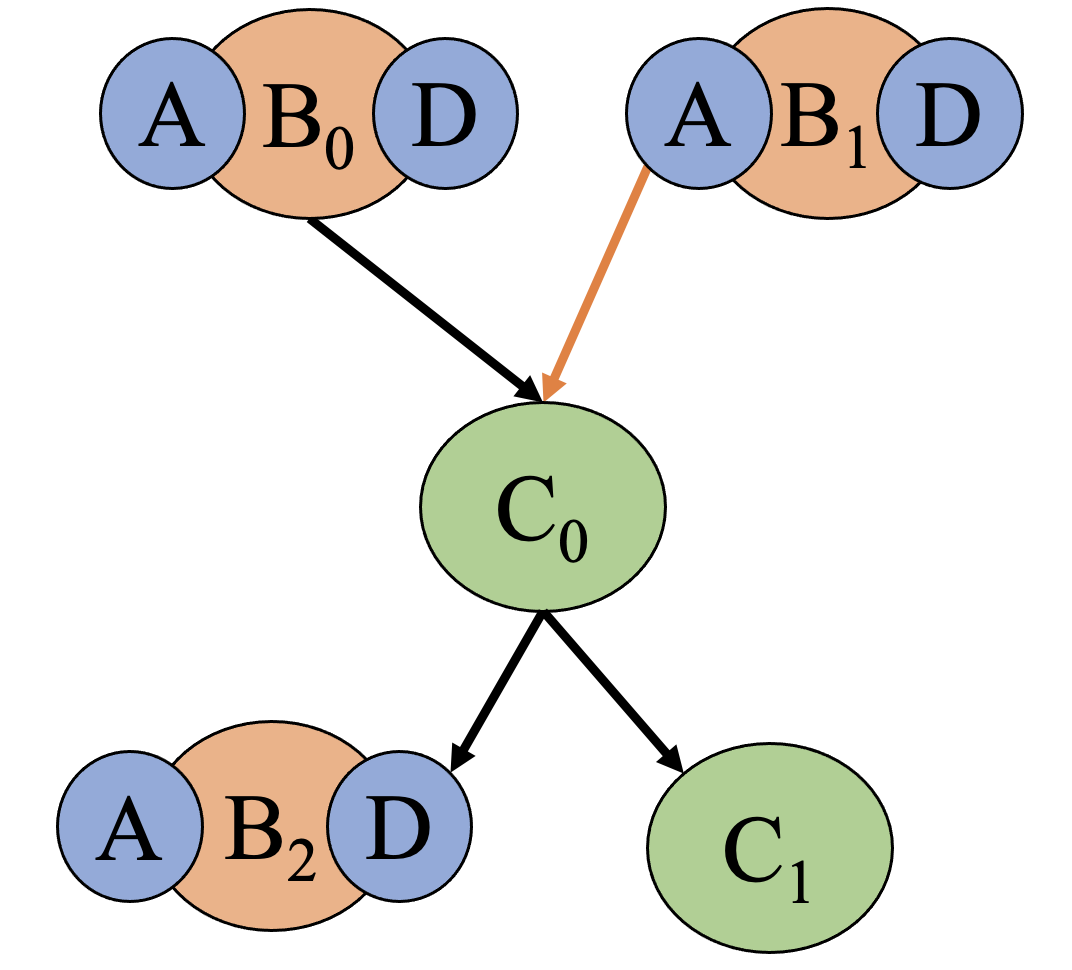}
     \caption{Illustration of SV types and relationships. The figure shows BSVs ($B_i$), their DSVs for activation (A) and deactivation (D), and CSVs ($C_i$). Here, CSV $C_0$ takes as positive source BSV $B_0$, as negative source the activation DSV of $B_1$; and conditions the CSV $C_1$ as well as the deactivation of $B_2$, modelling "$B_2$ is deactivated and $C_1$ is active if $B_0$ is active and $B_1$ is not activated."}
     \label{fig:svs}
 \end{figure}

BSVs are essentially environment observations, while DSVs represent their changes.\footnote{In our implementation we also use BSVs to represent actions taken by the agent in the previous step. Differently from environment observations, the actions do not have associated DSVs, since their activation and deactivation is in agent's control.} CSVs model the presence or absence of a relationship between a learned condition (sources) and its effect (active target states), indicated by the CSV being active (1) or inactive (-1). Figure \ref{fig:svs} shows these SV types and their connections. Note that CSVs are \textit{not} feedforward computational units; they represent the relationship between sources and targets - states of their targets are set independently of the CSV, unlike feedforward units that determine target states based on source states. CSVs partially function as feedforward units only when used for prediction of alternative outcomes.

\begin{figure}
     \centering
     \begin{subfigure}[t]{0.22\textwidth}
         \centering
         \includegraphics[width=\textwidth]{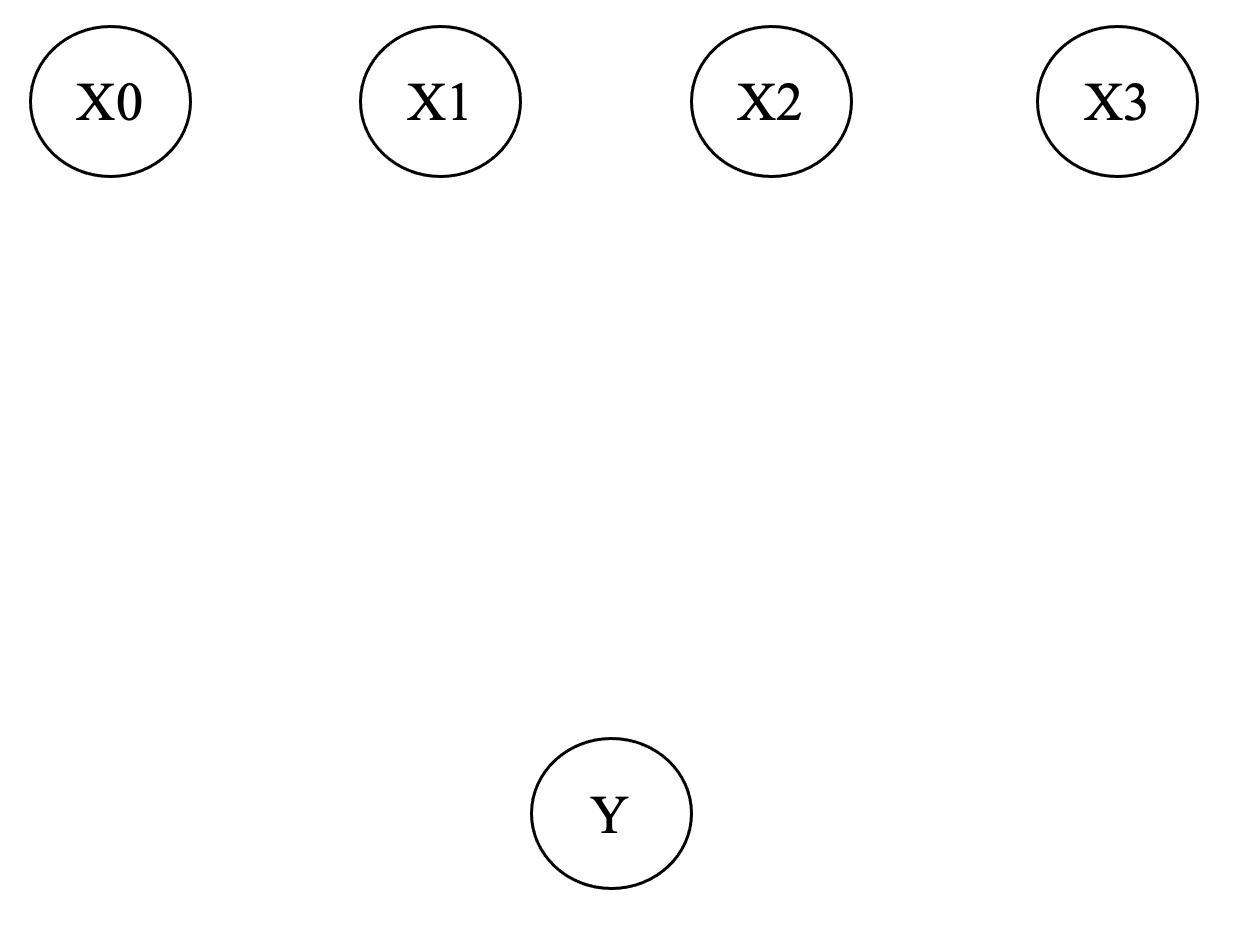}
         \caption{}
         \label{fig:csvform_1}
     \end{subfigure}
     \hfill
     \begin{subfigure}[t]{0.22\textwidth}
         \centering
         \includegraphics[width=\textwidth]{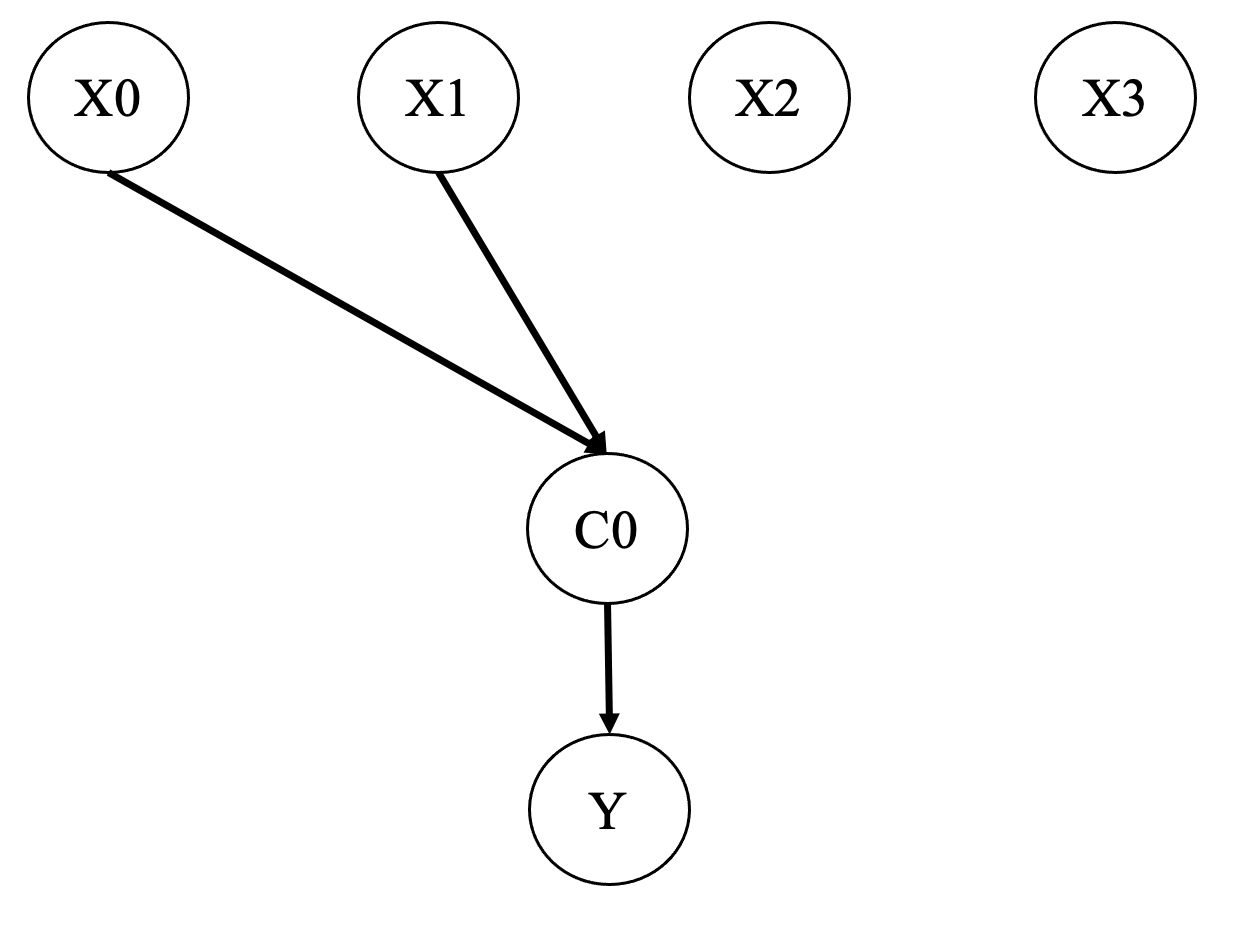}
         \caption{}
         \label{fig:csvform_2}
     \end{subfigure}
     \hfill
     \begin{subfigure}[t]{0.22\textwidth}
         \centering
         \includegraphics[width=\textwidth]{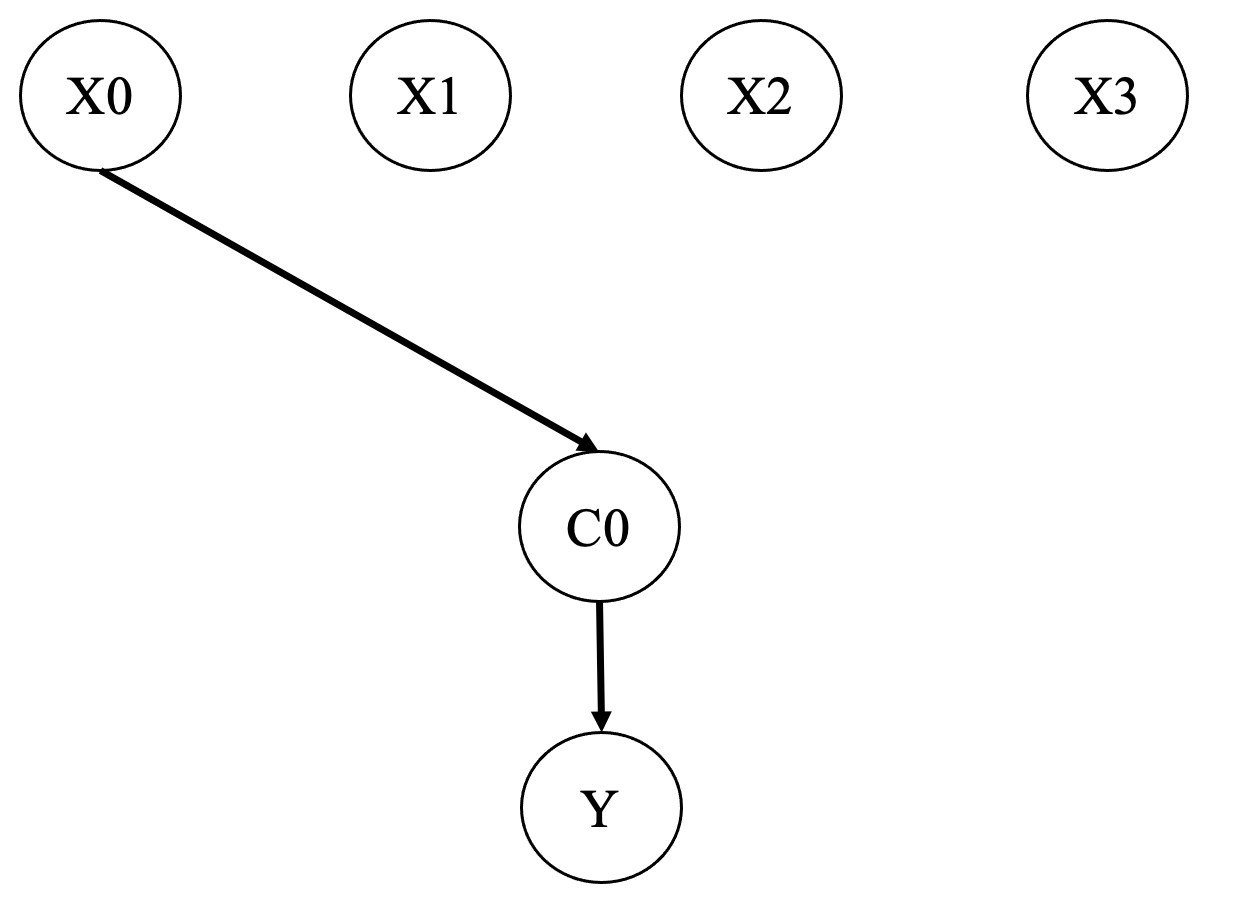}
         \caption{}
         \label{fig:csvform_3}
     \end{subfigure}
     \hfill
     \begin{subfigure}[t]{0.22\textwidth}
         \centering
         \includegraphics[width=\textwidth]{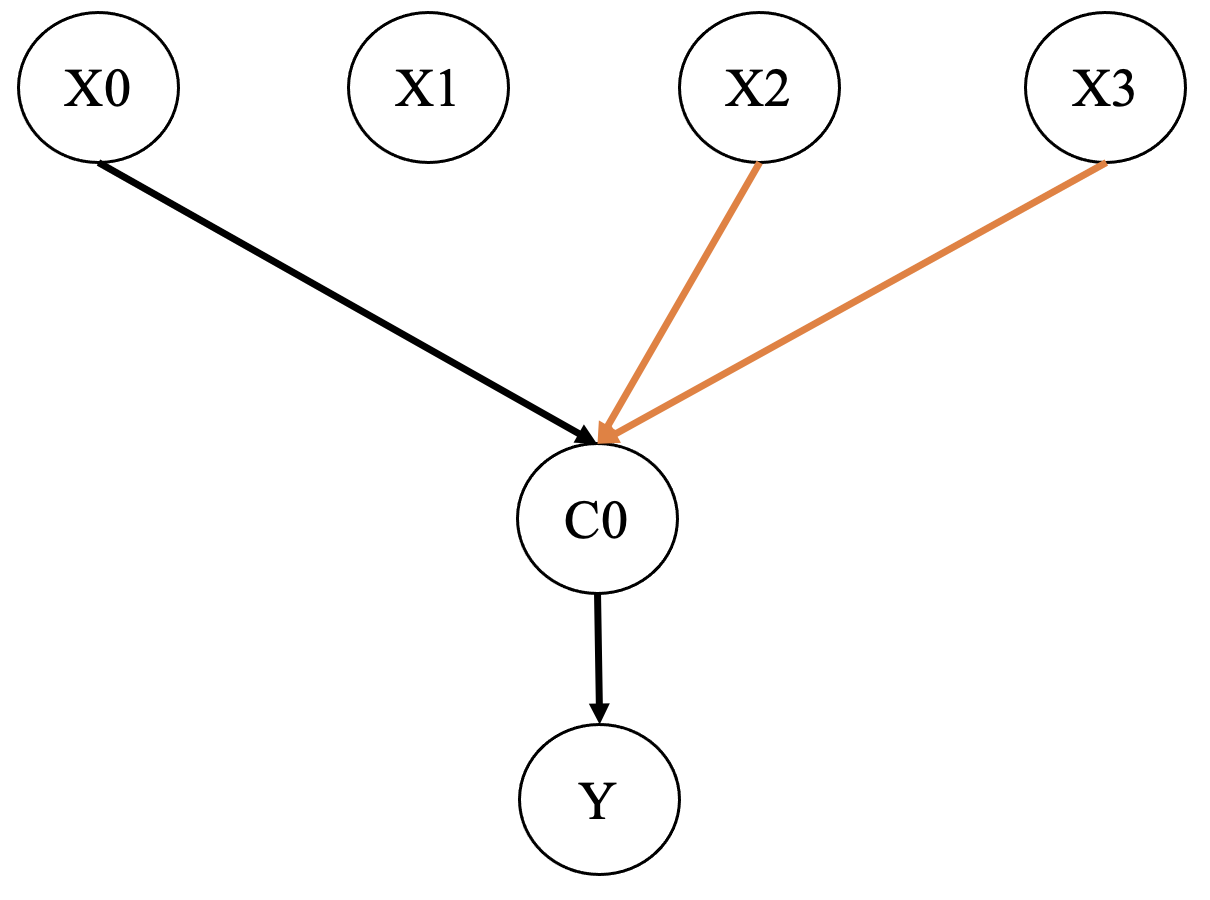}
         \caption{}
         \label{fig:csvform_4}
     \end{subfigure}
     \hfill
     \begin{subfigure}[t]{0.22\textwidth}
         \centering
         \includegraphics[width=\textwidth]{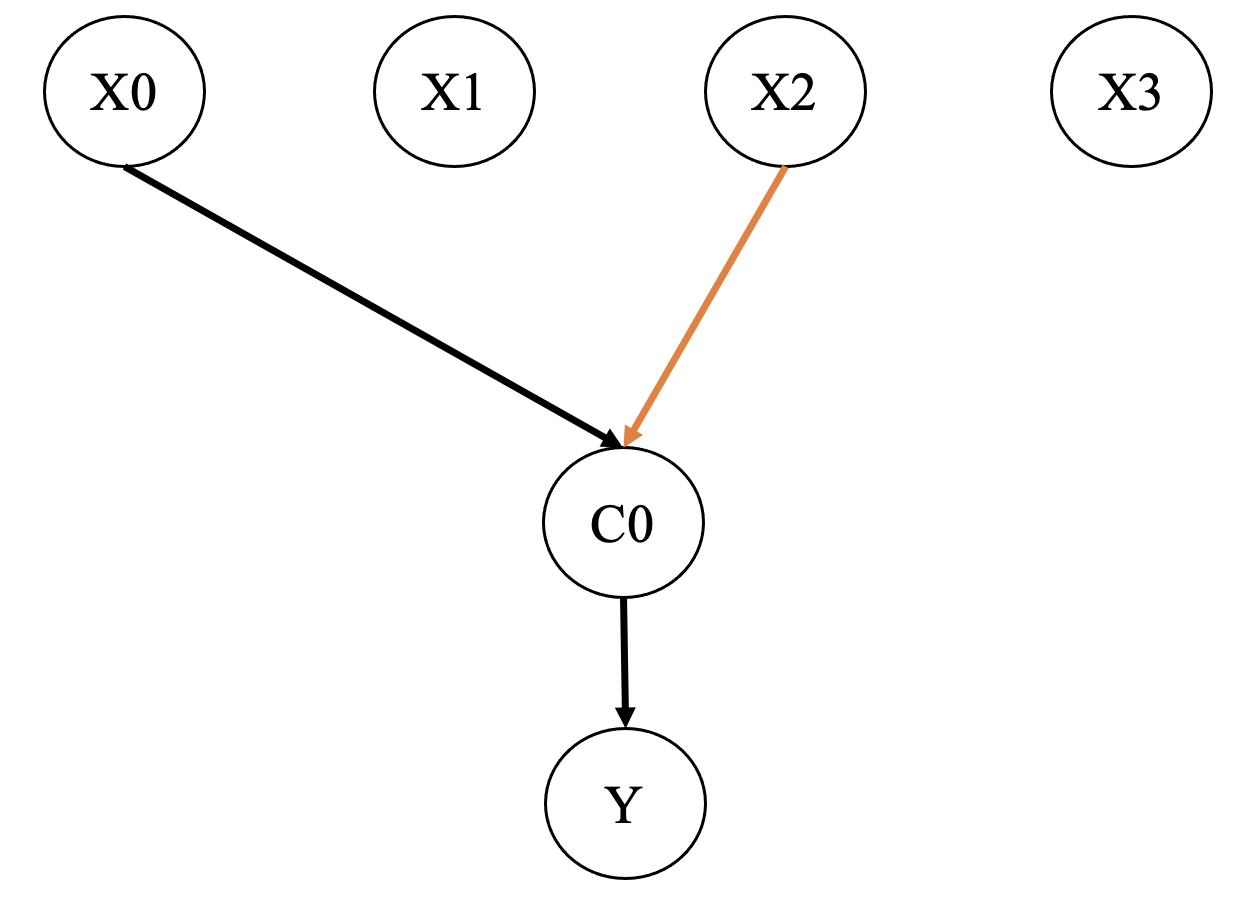}
         \caption{}
         \label{fig:csvform_5}
     \end{subfigure}
    \caption{Sample formation of a CSV in a continual manner. The relationship to be modelled is $Y = X0\ and\ !X2$ ("!" denotes "not"). Black and orange arrows represent positive and negative sources for CSV $C0$ respctively. $Xi$ can be interpreted either as single or grouped SVs. (a) Initial state with no relation formed between $X0-3$ and $Y$. (b) $X0, X1 \rightarrow Y$ observed. Positive connections hypothesizing both $X0$ \& $X1$ are required for Y are formed. (c) $X0 \rightarrow Y$ is observed. $X1$ is deduced unnecessary for $Y$. (d) $X0, X2, X3 \rightarrow !Y$ observed. $Y$ is hypothesized to be suppressed by $X2$ and $X3$. (e) $X0, X2 \rightarrow !Y$ observed. $X3$, seen unnecessary for suppression of $Y$, refined. Correct structure learned and is stable from now on.}
    \label{fig:csvform}
\end{figure}

The learning process proceeds \textit{step-by-step}, without incorporating an aggregate evaluation of multiple observational samples gathered from the environment, nor relying on iterative, repeated passes over a batch of data—distinct from traditional approaches such as deep RL. Initially, the model includes only BSVs and their DSVs, with no CSVs. At each step, Modelleyen seeks to explain the observed states of CSVs and DSVs in the previous timestep (modeling BSVs indirectly via DSVs). It does so by creating new CSVs to account for unexplained DSVs and CSVs. These retrospective explanations captured by CSVs become predictions for potential outcomes in the next timestep. Learning capability of Modelleyen comes from the \textit{operations} on CSVs - their formation, and the modification of their positive and negative sources; summarized as follows (detailed on Algorithms \ref{alg:algorithm_adaptationloop} and \ref{alg:algorithm_csvstate}):

\textit{Initial formation:} Figure \ref{fig:csvform_2}. At each step, if there are active DSVs or CSVs without an explanation (an active conditioner or an unconditionality flag, see Appendix), a new CSV is generated to explain them. Initially, the CSV has no negative sources ($X_N = \emptyset$) and includes all active BSVs and DSVs at that timestep as positive sources ($X_P$). No additional positive sources can be added to the CSV.

\textit{Negative connections formation:} Figure \ref{fig:csvform_4}. At the first instance where a CSV's sources are satisfied but its state is inactive, the CSV receives all active DSVs and BSVs at that timestep as negative sources ($X_N$), similar to previous step. No additional negative sources are added thereafter.\footnote{This process is separate from initial sources' formation to avoid creating exhaustive negative connections where unnecessary. Otherwise, a negative connection would be made with everything inactive during CSV creation, which, while accurate, would be overly exhaustive and unnecessary for most negative sources.}

\textit{Refinements:} Figures \ref{fig:csvform_3} and \ref{fig:csvform_5}. When a CSV's state is determined as 1 with at least one active positive source and active targets, we remove nonactive positive sources (${x \in X_P : S_X \neq 1}$) from $X_P$ and active negative sources (${x \in X_N : S_X = 1}$) from $X_N$. When the state is 0, with at least one active positive source, inactive targets, and at least one active negative source, we remove nonactive negative sources (${x \in X_N : S_X \neq 1}$) from $X_N$.

Intuitively, a CSV starts by being connected to all active SVs at formation, representing a comprehensive hypothesis of relationships. These relationships are then refined based on observations where some connections are deemed unnecessary, ensuring the CSV remains consistent with past observations locally. This refinement is central to Modelleyen's continual learning ability, evident from its lowest organizational level of CSVs, as formalized of the following property.

\begin{theorem}
    Let $y_i$ be an \textit{instance} that includes the previous states of all the positive and negative sources of a CSV $C$ and the current states of all its conditioning targets. Then, if $C$ undergoes any modification as a result of encounter with an instance $y_1$, its state in reponse to any past instance $y_0$ is not altered by this modification; as long as its set of targets remain identical and $C$ does not undergo negative sources formation (either because inactive state is not observed or because it has already undergone it).\footnote{The requirement for identicality of targets in this theorem is only to account for the fact that heterogeneous targets result in duplication of CSVs - see the Appendix for details of this mechanism. The theorem holds when one considers the response of the duplicated CSVs with respect to the targets assigned to each duplicate as well.} For the proof, see Appendix \ref{sec:app_proof}.
\end{theorem}

Theorem 1 is exemplified in Figure \ref{fig:csvform}: In \ref{fig:csvform_2}, after elimination of $X1$ as a positive source, the earlier exposure of $X0, X1 \rightarrow Y$ still results in a state of activity in $C0$, and likewise for $X2$ \& $X3$. With this property, we know that the state of a CSV in response to any past encounter is not altered except possibly for initial negative sources formation (happening only once per CSV), hence realizing continual learning without destructive adaptation in Modelleyen inherently and from the lowest level of organization.

\begin{figure*}
     \centering
     \begin{subfigure}[t]{0.25\textwidth}
         \centering
         \includegraphics[width=\textwidth]{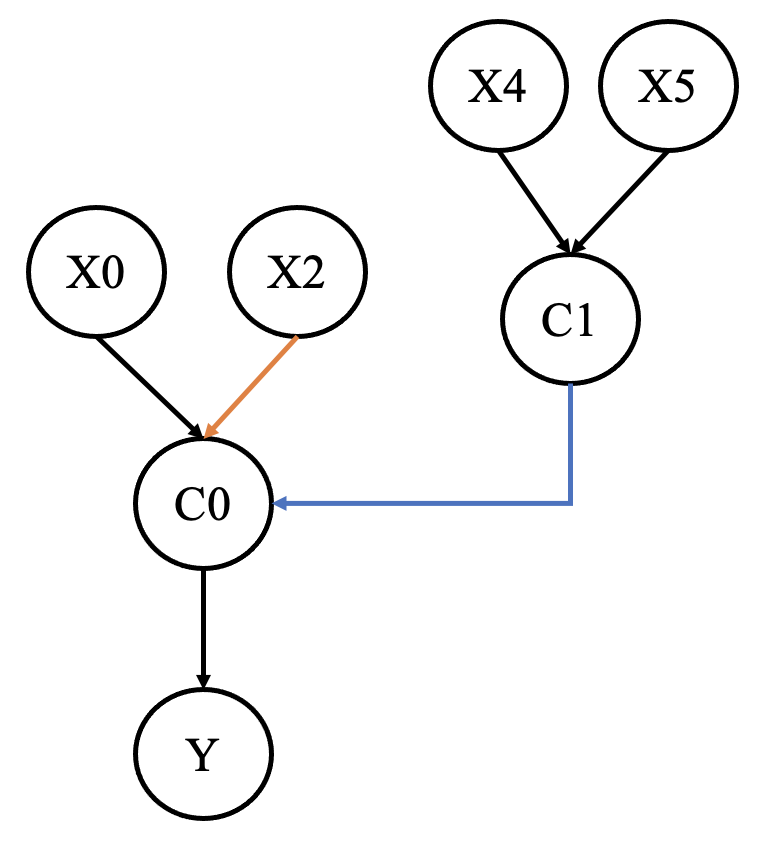}
         \caption{}
         \label{fig:csvformupstream_1}
     \end{subfigure}
     \vline
     \begin{subfigure}[t]{0.5\textwidth}
         \centering
         \includegraphics[width=\textwidth]{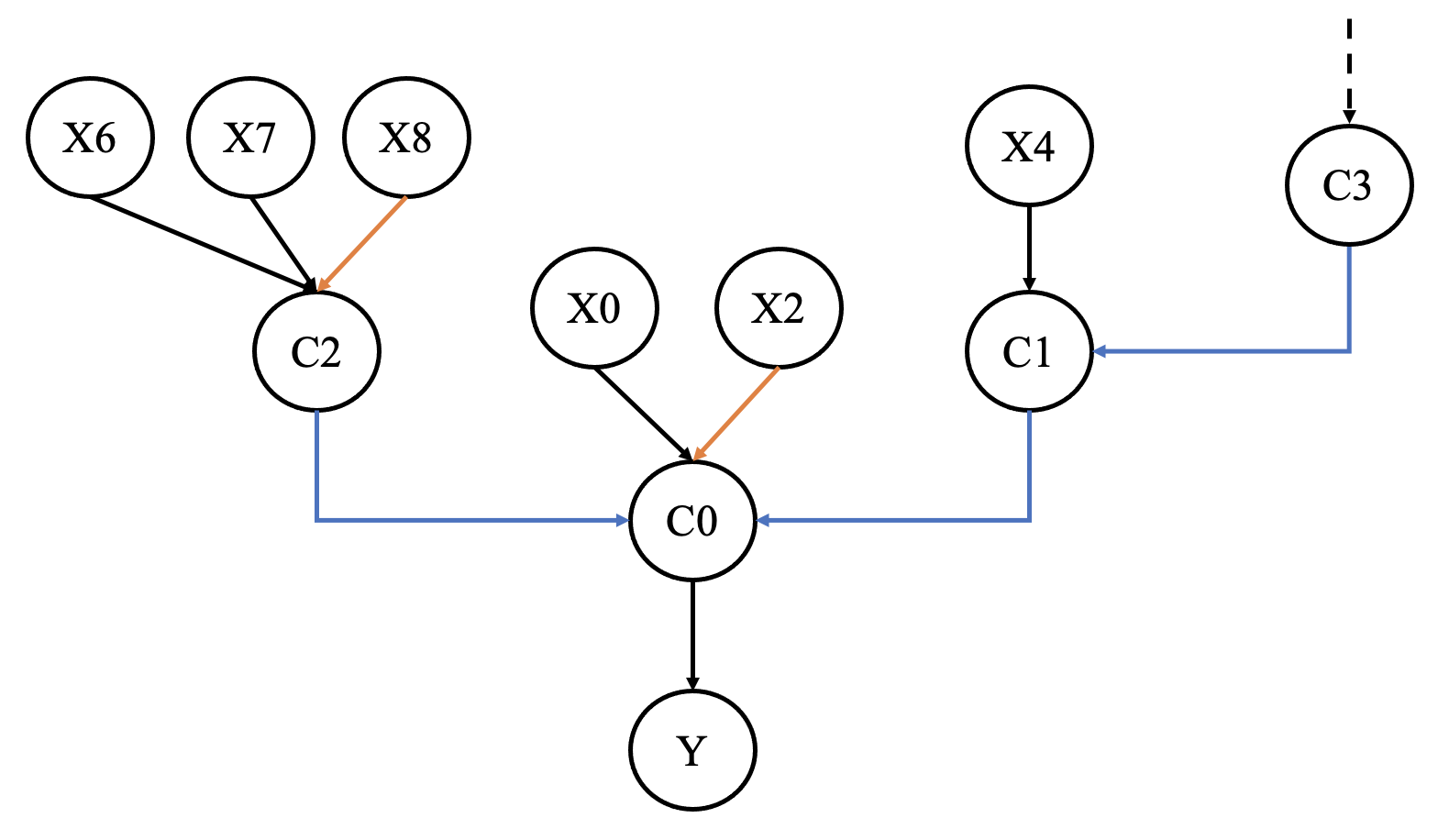}
         \caption{}
         \label{fig:csvformupstream_2}
     \end{subfigure}
    \caption{Example of upstream conditioning, continuing from Figure \ref{fig:csvform}. Assume that the unconditionality flag of $C0$ is set following an observation that $(X0,\ !X2)$ did not result in its activation (see main text). (a) $X0, !X2, X4, X5 \rightarrow Y$ observed. $C0$ is observed to be active, since $XO, !X2$ led to $Y$. A new CSV $C1$ is formed \& conditions $C0$. Note that $(X4, X5)$ alone will not predict activation of $C0$ if $C0$’s sources are not also active. (b) New conditioners are also subject to the CSV processes: Here, the source $X5$ of $C1$ has been refined, and new conditioners $C2$ and $C3$ are formed. Multiple conditioners represent alternative paths: In this case, $C0$ is expected to be active when sources of either $C1$ or $C2$ is active. Any logical function can hence be incorporated in a conditioning pathway in a minimal and ongoing manner without destroying past knowledge.}
    \label{fig:csvformupstream}
\end{figure*}

A CSV can condition/predict not only the activation of direct environmental dynamics (DSVs), but also possibly the activation of other CSVs. This capability enables the model to become more complex upstream, allowing for the representation of arbitrarily complex logical relations in a structurally minimal way, without requiring any \textit{a priori} knowledge of the existence of such relations. (As a result, it is not constrained by the assumptions such as those in \cite{mordoch2023learning} discussed in the introduction). This formation of upstream conditioning pathways is exemplified on Figure \ref{fig:csvformupstream}, continuing our example from Figure \ref{fig:csvform}. The processes of refinements, negative sources formations, and even further upstream conditioning are identical regardless of what the target of a CSV is.

Additionally, we quantify the statistical significance of relationships between each CSV and their targets - this prevents excessively large models and instability in environments with numerous observations and spurious relationships, expected to be especially important when scaling to higher-dimensional environments. For this purpose, we use a straightforward metric we called \textit{normalized causal effect (NCE)}, quantifying the amount of increase in probability of incidence of $T$ ($I(T)$) that satisfaction of sources of CSV $C$ ($SS(C)$) causes, normalized by the original probability of incidence:

\begin{equation}
    NCE = \frac{P(I(T)|SS(C)) - P(I(T))}{P(I(T))}
\end{equation}

Details \& reasoning behind this mechanism of statistical significance tracking are excluded from the main text for brevity can be found in Appendix \ref{sec:statistical_significance}.


This learning approach is fundamentally different from methods like NNs. In Modelleyen, the agent updates its model instantly with new information at each step, unlike other methods that make incremental adjustments over many steps. This process can be seen as the agent initially "overfitting" to observations —fully accounting for them— while gradually refining the model to be as structurally and explanatorily minimal as possible without contradicting past experiences. At every stage, the model is as general as necessary based on prior exposures, but no more. The more specific representation (e.g., more sources per CSV) allows for precise generalization when new observations arise, increasing likelihood of consistency as sources are refined. This mechanism is central to Modelleyen's continual learning capability and reflects a fundamental process in biological systems, where redundant variations are maintained and selected as needed \cite{gerhart2007theory}. Without knowledge of any prior approach grounded in these principles, we propose naming such learning mechanisms —which rely on local, component-level variation and selection as exemplified by Modelleyen— as \textit{varsel mechanisms}; and networks constructed using them as \textit{varsel networks.} Unlike conventional methods that start with underfitting and progressively adjust while avoiding overfitting, this concern is irrelevant in varsel networks, as the necessary level of generalization is inherently built into the model based on all previous observations.

\section{Planlayan}
\label{sec:planlayan}

We introduce \textit{Planlayan}, an extension on Modelleyen designed to demonstrate goal-directed planning through backward tracking from desired goal states to current states.

\textbf{\textit{Preprocessing the model and Group SVs:}} We first briefly preprocess a learned model to reduce the number of connections. To this end, we group the sets of BSVs in our that are either (1) collectively act as positive or negative source of a CSV, or (2) have an event that is collectively predicted by a CSV. Each such grouping becomes a \textit{constituent} of a Group SV (GSV). For example, if a CSV $C0$ has positive sources $(B0, B1, B2)$ and predicts deactivation of $(B3, B4)$; then two GSVs are created: $G0 = (B0, B1, B2)$, $G2 = (B3, B4)$. This preprocessing stage is only for practical purposes and is not in principle needed for the operation of Planlayan, but we think it is essential for scalable representations of models learned by Modelleyen in the long run.

\textbf{\textit{Main Process of Planlayan:}} Planlayan constructs an action network (AN) based on a model generated by Modelleyen, incorporating alternative outcomes. An AN is a dependency graph with root nodes representing the current environmental states (current BSV, GSV, and DSVs), along with possible alternative connections (shown by multiple conditioning links from CSVs) needed to achieve a specified goal state variable (see Figure \ref{fig:fullAN} example from experiments). To build this, we use a simple recursive function that generates the upstream action network for a given node (Figure \ref{fig:angen} - see Algorithm \ref{alg:planner} in Appendix for details). At each call, the function adds predecessors for the specified node until it reaches the root nodes that represent current environmental states. These predecessors vary by state variable types based on their model functionality, as summarized in Figure \ref{fig:angen2}.

\begin{figure*}
     \centering
     \begin{subfigure}[t]{0.25\textwidth}
         \centering
         \includegraphics[width=\textwidth]{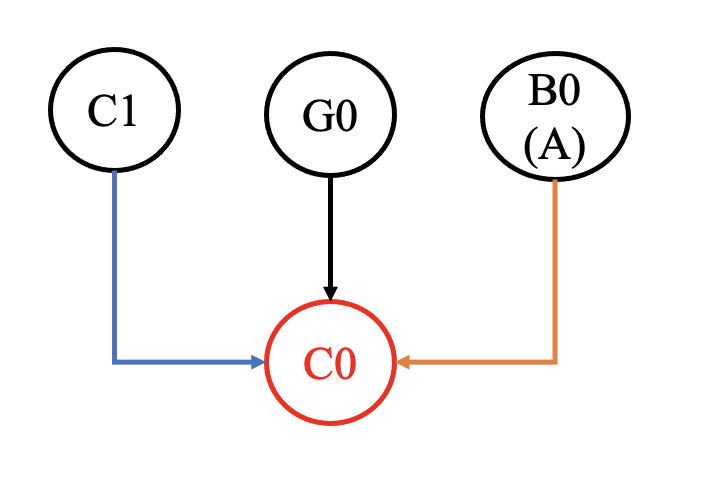}
         \caption{}
         \label{fig:angen1}
     \end{subfigure}
     \vline
     \begin{subfigure}[t]{0.5\textwidth}
         \centering
         \includegraphics[width=\textwidth]{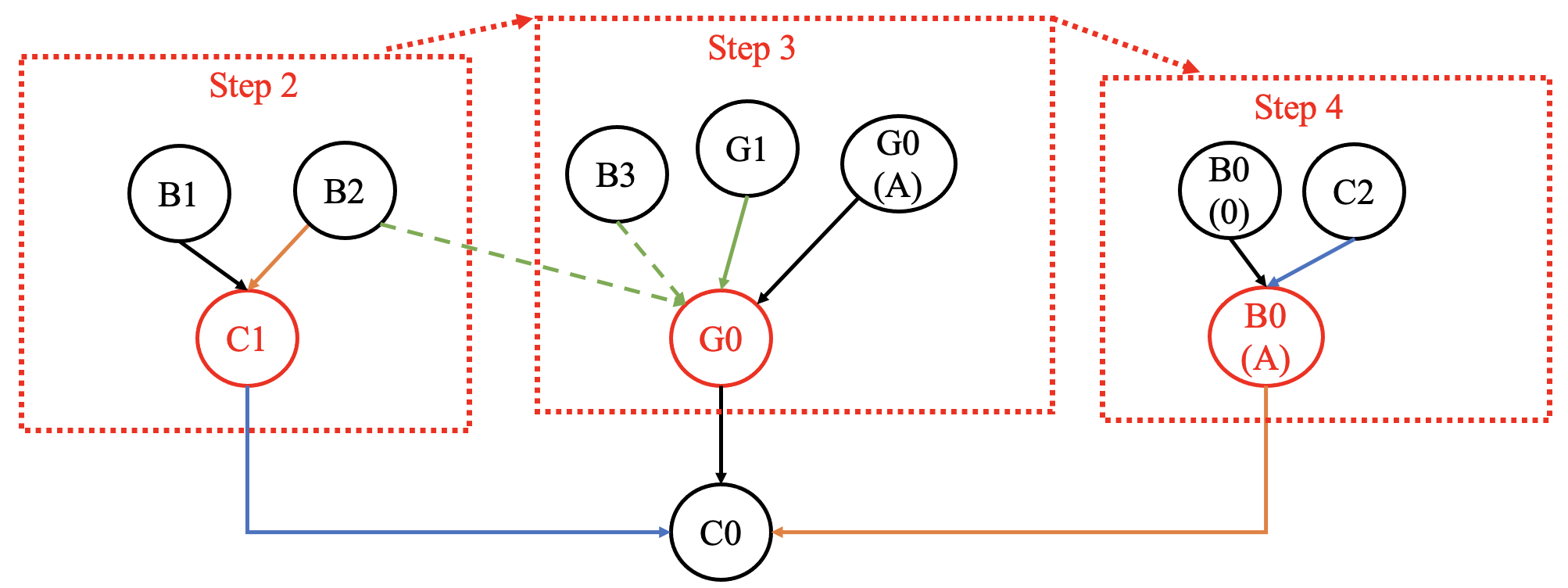}
         \caption{}
         \label{fig:angen2}
     \end{subfigure}
    \caption{Illustration step-by-step upstream generation of action network, operating on different SV types. BX, CX and GX stand for BSV, CSV and GSV nodes respectively, (A) for activation, (0) for nonactive state. Black arrows are positive sources and precondition targets, green arrows are constituent (dashed) and constituency (solid) relations. The node that is extended at each step is highlighted in red. (a) Step 1. CSV C0 is opened. For CSVs, their upstream conditioners (C1) and sources are expanded (G0, B0(A)). (b) Steps 2-4. Each step opens up one of the sources of previous step. For GSVs (G0), constituents (B2, B3), constituencies (G1) and precondition events (G0(A)) are opened. For DSVs (B0(A)), their precondition states (B0(0)) and their conditioners (C2) are opened. Possible interrelations (e.g. B2 for C1, G0) do not need reopening if they already exist.}
    \label{fig:angen}
\end{figure*}

\textbf{\textit{Action Choice:}} The agent generates an action network each time it needs to select an action. (While this is computationally unnecessary—since the agent could reuse a generated AN until it reaches the goal by tracking its position along the AN—we maintain this approach for simplicity.) From the generated AN, the agent identifies actions that can immediately activate any CSV in the action model, specifically those whose sources and sources of their downstream targets do not involve any unactualized BSV states. The agent then randomly selects one of these actions for the current step. Since only one action is chosen, the agent can consider the entire AN including alternative pathways.

Planlayan is explicitly goal-directed, identifying a path from initial states to the goal without needing rewards, although rewards can help prioritize the search. Unlike methods like model-based RL, which typically search from initial states to goals via forward-sampling, Planlayan considers both initial and goal states, focusing on steps derived from the environment model. The planning algorithm is a simple search method that unfolds upstream action networks from the model, as our main aim is to demonstrate the interface between Modelleyen’s modeling components and general deliberative behavior without going into extensive detail. Planning is a well-established field with efficient methods and useful heuristics \cite{ghallab2016automated}, and once the interface between Modelleyen and Planlayan is established, implementing more advanced algorithms is straightforward.

Finally, we note two visible limitations of the current version of Planlayan. First, the generated action networks are exhaustive, including every possible path to initial states. Second, the current version does not account for the precise timing of multiple events. In our experiments, for instance, the RS environment subtype (see Figure \ref{fig:environment}) takes longer due to the BSV \textit{DO} having two pathways for deactivation, the correct one being the one that deactivates BSV \textit{W} as well at the same time. The planner fails to distinguish between these pathways, leading to some unnecessary loops. These limitations are not addressed in current framework to keep its simplicity, since they do not affect our demonstrative use of Planlayan to a major degree. They are discussed in Section \ref{sec:conclusion}.

\subsection{Overview of the Agent's Operation Flow}

In summary, the operation of an agent utilizing Modelleyen and Planlayan follows these steps, repeated continuously as the agent interacts with the environment in an online manner, without the need for episode division or offline learning periods:

\begin{enumerate}
    \item Execute actions and gather the resulting observations from the environment.
    \item Process the environment's observations and update the model (Modelleyen - Section \ref{sec:Modelleyen}, Algorithms \ref{alg:algorithm_adaptationloop} and \ref{alg:algorithm_csvstate}.)
    \item Generate a plan based on the current model and goals, then select an action from the resulting plan (Planlayan - Section \ref{sec:planlayan}, Algorithm \ref{alg:planner}.)
\end{enumerate}

\section{Behavior Encapsulation}
\label{sec:behavior_encaps}

Modelleyen and Planlayan together create a complete system capable of continual learning and structured goal-directed behavior. However, the exhaustive action networks produced by Planlayan do not exemplify a comprehensible representation, which is one of our key goals. Additionally, Planlayan does not fully leverage this structured representation to address a long-standing challenge in AI behavior learning: the decomposition of learned behavior into subunits defined by automatically determined preconditions and consequences in an arbitrary hierarchical manner. To address this, we introduce a behavior encapsulation mechanism that operates on the action networks generated by Planlayan, transforming flat, exhaustive action plans into a hierarchically structured and comprehensible format.

\begin{figure}[t]{}
     \centering
     \includegraphics[width=0.5\textwidth]{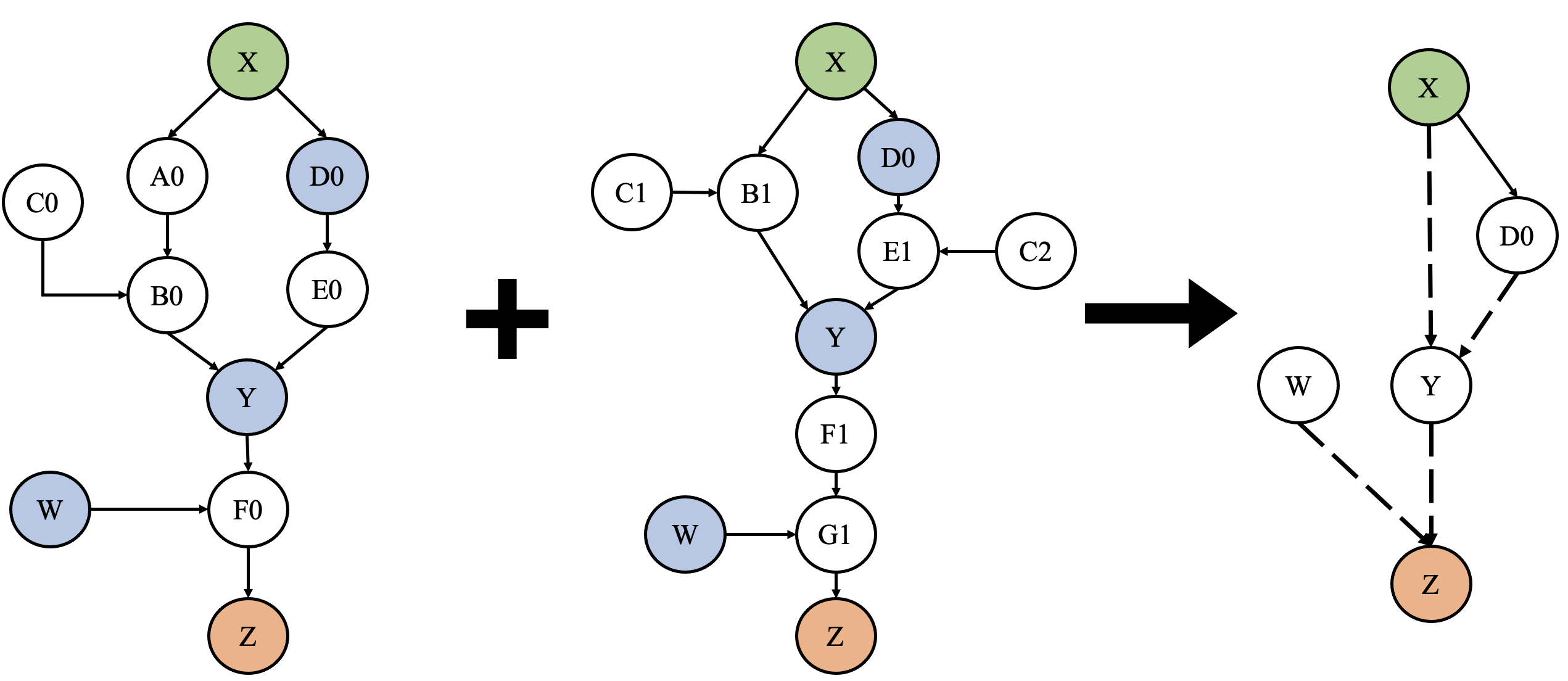}
     \caption{Illustrative example for the aim of behavior encapsulation process. To the left are two action networks (ANs) that represent two alternative pathways, split from the unified AN generated by Planner (node names are placeholders and can be of any SV type and target effect). We want to encapsulate the pathways between X and Z. For that; all pathways that are reliably present in all (here, both) networks are identified and a new \textit{encapsulated AN (EAN)} is formed with them (right). Each encapsulated edge (dashed) in EAN includes copies of subnetworks that corresponded to this pathway in the original AN variants; which can be further encapsulated in subgroups via a recursive call (for example, edge (D0,Y) would include two pathways; first one formed only of E0, the second of C2 and E1). The EAN on right can be regarded as the \textit{subpolicy} for realization of Z from X.}
     \label{fig:encapsulation_example}
 \end{figure}

The action network (AN) produced by Planlayan contains multiple alternative pathways. Our first step is to isolate each pathway into individual alternative action networks by creating copies of the original network, each including only one of the conditioning alternatives for each CSV and DSV. Next, we aim to develop a reduced, high-level network that captures the reliably observed pathways across all these alternative ANs (see Figure \ref{fig:encapsulation_example} for an abstract example, and Figure \ref{fig:encapsulatedAN} for a specific case from our experiments). The nodes in this new graph represent necessary subgoals for the current goal, while the encapsulated edges denote the subpolicies linking their start and end states. We achieve this through a simple, edge-oriented process that starts with one action network and refines edges by removing those whose source and target aren't connected in other ANs, while linking all relevant predecessors and successors. This process continues until no further changes occur, resulting in a minimally structured version.

After generating the high-level network, we isolate the subgraphs that connect the subgoal nodes, representing them as alternative pathways for the corresponding subpolicies. This process is done recursively on the internal encapsulated subnetworks by grouping networks that share at least one common node, continuing until no such groups can be formed. This results in a behavior representation that, while complex in its extended form, is maximally structured and comprehensible at each organizational level. Although this process is computationally intensive, it only needs to be executed once for each action path, as long as the underlying model remains unchanged, making the computational complexity manageable.

Beyond enhancing the comprehensibility of action networks post-hoc, this encapsulation process can significantly aid agent behavior. Encapsulated behavioral subunits, (whose delimiters are not provided to the agent in advance), can be reused when the same precondition/goal pairs arise. We do not yet perform this integration of behavior encapsulation with the agent's ongoing operations, and present it separately as an illustration of what becomes possible with AAI.

\section{Experimental setup}

\begin{figure}[t]{}
     \centering
     \includegraphics[width=0.5\textwidth]{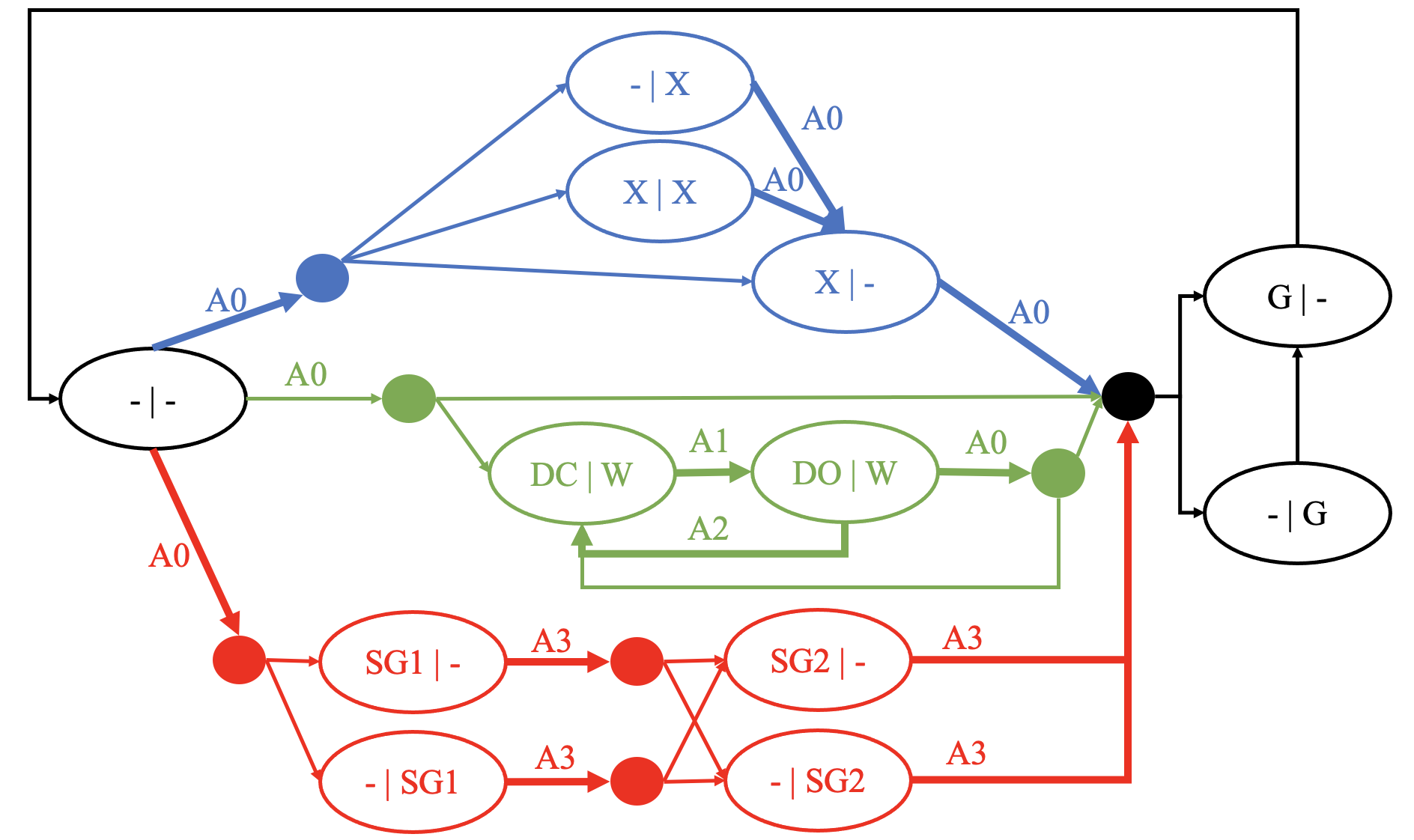}
     \caption{The environment and its subenvironments that we test on, essentially a FSM with two cells each of whom can take one of the states "DO, DC, W, G, SG1, SG2, X" or be empty ("-"). Each state is connected with arrows representing succession relations between them; filled circles correspond to multiple alternatives that can result from it. Green, red and blue portions are "RS", "SG", and "NEG" subtypes respectively (detailed in text), black portion is included in all subtypes. In "Complete" variant, all transitions and states are included. The agent’s goal is to activate state "G" in the first cell, and optimal actions are indicated by bold transitions. The environment has 20 actions, much larger than what is actually useful, in order to make it difficult to reach goal randomly.}
     \label{fig:environment}
 \end{figure}

We demonstrate the operation of AAI on a simple test environment, which is a finite-state machine (FSM) with two cells, each capable of seven states or inactivity, as shown in Figure \ref{fig:environment}. The environment includes three subtypes ("RS", "SG", "NEG"), illustrated by different colors. This setup was designed to model various types of temporal successions, such as basic succession, correlated changes, alternative causes/outcomes, uncertain transitions, and negative conditons.\footnote{The environment was vaguely inspired from Multiroom environment in Minigrid \cite{MinigridMiniworld23}. For intuition behind this FSM, see the Appendix.} There is also a \textit{random} variant of the environment where two additional states that get activated randomly are introduced, in order to test statistical significance filtering mechanisms. This environment was chosen in order to validate the core operation of AAI in a simple and understandable setting, which made in-depth analysis and debug of the design very feasible during development process. There is no inherent limitation to applying to more complex environments, akin to those used for testing e.g. RL algorithms,\footnote{With the possible exception of high-dimensional visual inputs, which will need an extension of AAI to incorporate their inherent structure, akin to Convolutional NNs as compared to fully connected ones.} except that the planner implementation should incorporate the changes needed to make search nonexhaustive (see Sections \ref{sec:planlayan} and \ref{sec:conclusion}). We leave validation on such environments and changes in design to future work, as this presentation is dense enough already.

In our base planning experiments, we compare the performance of an agent that learns a model followed by planning (with a 10\% chance of random actions for exploration) to one that acts purely randomly. The agent starts with 4000 random actions to learn the environment model, then uses Planlayan for the next 4000 steps. We measure the average steps to reach the goal before and after planning. Next, we conduct continual learning experiments where the agent learns with predefined goals and the environment subtypes switch every 500 steps (with readaptation) or 1000 steps (without readaptation). We test whether the agent can achieve similar performance in different subtypes, both in vanilla and random environment variants without any readaptation of the model, and also analyse learning progression when readaptation is enabled. Finally, we present a demonstrative case of behavior encapsulation on a learned model. For more details on the experimental setup, see Appendix \ref{sec:appendix_expsetup}. We do not provide comparison with any existing method since we are not aware of any method that could provide a meaningful comparison: As discussed in Section 1, to the best of our knowledge, there are no existing methods in literature that can either perform unsupervised continual learning of an environment reliably with no task boundaries and no past sample replay, \textit{or} perform precise goal-directed behavior on a learned model together, \textit{or} encapsulate \& represent the behavior in an automatically generated arbitrarily hierarchical structure in a comprehensible manner, let alone solving all these seemingly disjoint issues with a common framework.

\section{Results and Discussion}

\begin{figure}[t]{}
     \centering
     \includegraphics[width=0.5\textwidth]{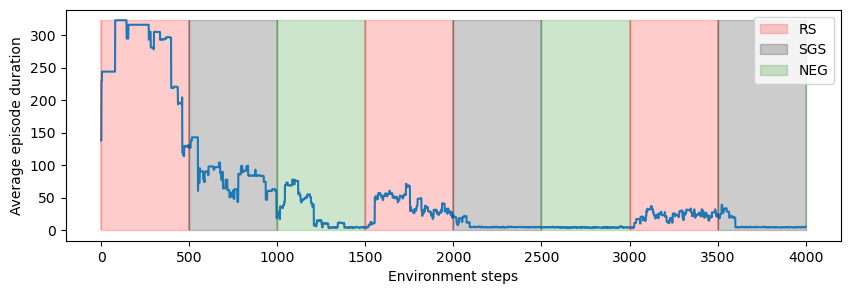}
     \caption{Average (5 trials) episode durations throughout learning with changing environment subtypes, with model readaptation enabled. Vertical limits show the environment changes, note that the actual step of change varies by a few steps across trials since end of the ongoing episode is waited.}
     \label{fig:CL_planning_nonrand_readapt}
 \end{figure}

\begin{figure*}
     \centering
     \begin{subfigure}[b]{0.3\textwidth}
         \centering
         \includegraphics[width=\textwidth]{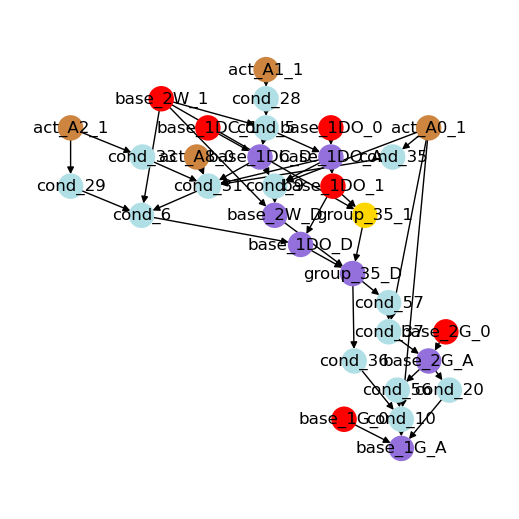}
         \caption{Full action network.}
         \label{fig:fullAN}
     \end{subfigure}
     \begin{subfigure}[b]{0.4\textwidth}
         \centering
         \includegraphics[width=\textwidth]{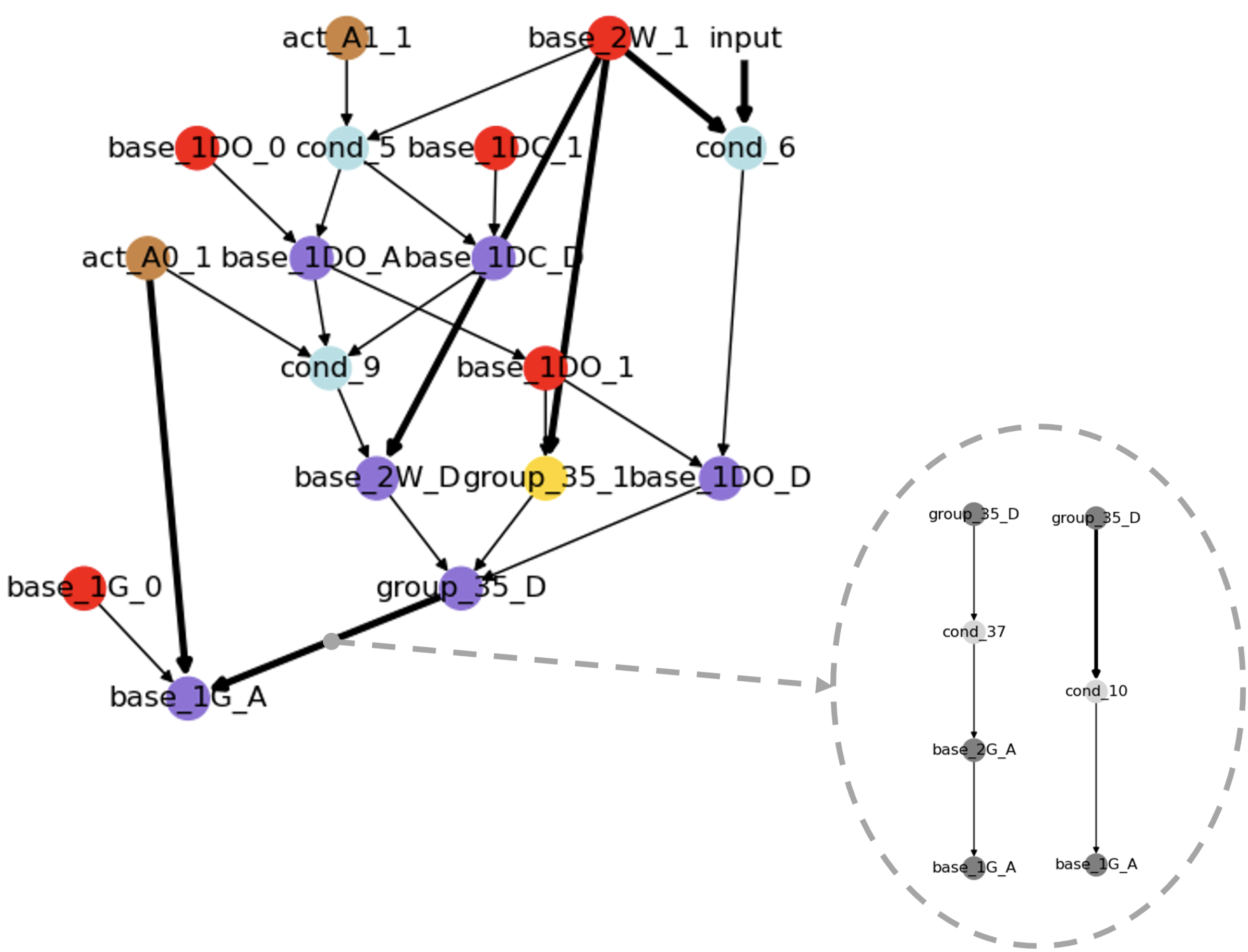}
         \caption{Encapsulated action network.}
         \label{fig:encapsulatedAN}
     \end{subfigure}
    \caption{Example of action networks on test environment. Bold edges are encapsulated. Each node represents a different state variable, and each edge represents conditioning and succession relations between them.}
    \label{fig:actionnets}
\end{figure*}

\begin{table}[]
    \centering
    \caption{Base goal-directed behavior. Mean episode durations (across 4000 steps) before the introduction of goal (no goal) and after it (with goal), for Complete (nonrandom) and Complete-Random variants of the environment. For the latter, Modelleyen's statistical significance filtering have been enabled. Actions are chosen randomly before the introduction of the goal. All results are averages across 5 independent trials. Inside paranthesis are standard deviations.}
    \begin{tabular}{c|c|c}
         & No goal & With goal  \\
        \hline
        Complete & 98.1 (17.69) & 7.28 (0.5) \\
        Complete-Random & 99.22 (32.61) & 22.33 (28.2) \\        
    \end{tabular}
    \label{tab:baseplan}
\end{table}

\begin{table*}[]
    \centering
    \caption{Continual learning. Mean episode durations with environment change, for vanilla, random environment, and readaptation variants. Columns represent the successive environment subtypes. Subtypes indexed "L" have model learning enabled, "NL" have it disabled (except for "readaptation" variant, which continues learning throughout the end). All results are averages across 5 trials.}
    \begin{tabular}{c|c|c|c|c|c}
         & RS-L & SGS-L & NEG-L & RS-NL & SGS-NL  \\
        \hline
        Vanilla & 45.58 (25.55) & 5.33 (0.28) & 4.47 (0.22) & 10.38 (1.68)  & 4.3 (0.11)\\
        Random Env. & 190.86 (148.0) & 32.3 (9.93) & 9.87 (3.45) & 121.69 (82.33) & 35.05 (5.42) \\        
        Readaptation & 89.01 (58.72) & 28.19 (21.45) & 6.06 (0.74) & 13.73 (3.45) & 4.71 (0.15) \\       
        \hline
        Random actions & 275.86 & 67.53 & 52.48 & 275.86 & 67.53 \\        
    \end{tabular}
    \label{tab:continual}
\end{table*}

\textit{\textbf{Base Planning:}} Table \ref{tab:baseplan} compares episode durations for random actions (first 4000 steps) and planning (next 4000 steps). The planner significantly reduces the time needed to reach goals compared to random actions. These results demonstrate AAI's effectiveness in accurately modeling the environment and performing goal-directed behavior. The agent consistently achieves similar performance across the 4000 steps after the goal introduction, indicating it can learn the environment independently of the goal and immediately realize the goal in a learned environment without further training. This efficiency reduces training costs compared to existing methods, as approaches like RL require a goal-dependent reward signal, necessitating some relearning when goals change, even in identical environments. However, randomness does have a notable impact: while planning and modeling remain effective, the presence of additional connections above the significance threshold leads to more redundant action choices. This issue arises from relying only on first-order significance and the challenge of establishing a universal causal effect limit, a limitation we will address in future work—see Appendix \ref{sec:statistical_significance} for details.

\textit{\textbf{Continual Learning:}} Table \ref{tab:continual} displays the agent's continual learning performance across changing environments, with the goal defined from the start. Both vanilla and random variants maintain or even improve their performance after exposure to different environments, often outperforming initial learning periods, without readaptation. For instance, the vanilla version averages 5.33 steps on the SGS variant during learning and 4.3 steps after intermittent exposure to other subtypes. Figure \ref{fig:CL_planning_nonrand_readapt} also illustrates this, showing that with model adaptation enabled, the agent performs consistently with its previous endpoint performance in the same environment subtype, without any spikes indicating destructive adaptation. Additionally, most steps are spent in the RS variant due to the precise timing requirements of Planlayan (as discussed in Section \ref{sec:planlayan}).

\textit{\textbf{Behavior encapsulation}} Figure \ref{fig:actionnets} shows a sample action network and a demonstration of the resulting encapsulated AN. Here the start states are (DC,W), hence encapsulation is between these states (and inactive states for all the rest) and the goal state. The full action network even for this simple environment is clearly very complex; however encapsulation can turn it into a comprehensible, structured, minimal format. On Figure \ref{fig:encapsulatedAN}, many paths that are seen to be alternatives have been encapsulated (example shown from \textit{Group35-D} to \textit{1G-A}), and only reliable (i.e. necessary) connections remain; which, upon inspection, can be seen to correspond to the transition (DC,W) $\rightarrow$ (DO,W) that is invariably needed for reaching the goal from (DC,W). As discussed before, the identified subgoals and pathways, as well as encapsulated components, can be used as building-block subpolicies for future behavior, though we did not yet incorporate this integration with ongoing agent behavior.

\section{Conclusion}
\label{sec:conclusion}

Agential AI, consisting of Modelleyen, Planlayan, and the behavior encapsulator, offers a promising solution to the core challenges in classical machine learning. This paper focuses on its strengths in continual learning, interpretability, the seamless integration of learning and planning, and the decomposition of behaviors into flexible hierarchies. The effectiveness of this approach stems from a shared foundation: constructing a structured model of the environment while retaining past knowledge through a method driven by local variation and selection. We coined the term \textit{varsel mechanism} to describe this class of learning methods, along with \textit{varsel networks} for the networks they construct, which adapt their structure through localized variation and selection, as exemplified by Modelleyen. We believe such methods, beyond the specific examples outlined here, hold significant potential to address the fundamental limitations of gradient-based machine learning.

The only inherent limitation of AAI is its reliance on discrete observation and state spaces. Addressing continuous spaces will require additional methods like preprocessing or analog-digital conversions \cite{pelgrom2013analog}. However, many relevant AI problems can be represented with non-continuous observations or converted into such formats (e.g., feature-based vision or tasks involving relative values). The primary exceptions are tasks that require precise, fine-tuned control; in such cases, AAI could work alongside statistical learning methods like neural networks for low-level behavior control. Therefore, explicit support for continuous spaces may not be necessary, as AAI is primarily designed for cognitive tasks in structured environments rather than control tasks.

\subsection{Future work}

As mentioned earlier, the current version of AAI serves as a foundation to demonstrate core mechanisms. It has some venues of development that will addressed in future work.

A primary class of issues that require attention revolves around the assumptions made regarding the structure of environment observations and how we model them. Specifically, the current model assumes a Markovian environment, focusing solely on immediate state transitions and neglecting long-term dependencies. Additionally, while Modelleyen can handle structured spaces like large visual observations (in the same way a fully-connected neural network processes images by representing each pixel as a separate feature), adapting it to more specific structures (like using CNNs or transformers instead of FCNNs) would enhance its scalability. Addressing those domains, along with potential others, requires adapting the Modelleyen algorithm to operate on networks as observations rather than lists of state variables. In particular, both visual spaces and temporal event chains can naturally be represented as finite networks with predefined structures. For visual spaces, this could mean networks of pixels or generalized base features such as edges; for event chains, the network would represent the sequence of events from the start of the current episode. These source networks could then undergo the same algorithmic steps presented in this paper (Figures \ref{fig:csvform} and \ref{fig:csvformupstream}) while maintaining the same guarantees of continual learning and minimality, with the primary modification being the redefinition of the refinement operation as "network refinement." This process, already discussed in Section \ref{sec:behavior_encaps} (Figure \ref{fig:encapsulation_example}) in the context of behavior encapsulation, would allow the method to represent visual or temporal spaces of any dimensions—or any other domain expressible as a network. Combined with Modelleyen’s ability to learn complex succession relationships, this extension would enable the algorithm to tackle tasks with high dimensionality.

In addition to expanding the types of observation spaces processed, our framework has a few additional avenues that will be explored in future work. First, the statistical significance calculations in Modelleyen currently focus only on first-order relationships. For a more precise tracking of significances, this should be extended to incorporate upstream conditionings. Additionally, to scale Planlayan for more complex environments, selective pathway extension during planning is required. This can be achieved using existing mechanisms in Modelleyen, such as returning immediately when a viable path is found or prioritizing pathways based on statistical significance. Precise timing in Planlayan, where needed, can also be managed by evaluating the full consequences of each pathway and excluding those that reverse precondition states or activate conditions that hinder future actions. Another direction for future development is incorporating behavior encapsulation into ongoing operations to enable reusable behavior patterns. This is a key motivation for behavior encapsulation, shared by related fields like hierarchical reinforcement learning. We believe that the structured representations learned by AAI offer an ideal foundation for this process.

Once these issues are addressed with future iterations, we believe the approaches we exemplify in this paper has the potential to significantly advance the development of more capable and controllable AI systems.

\newpage

.

\newpage

\appendix
\section{Appendix}

\subsection{Details of Modelleyen system components}
\label{sec:modelleyen_details}

We define a state variable (SV) as a variable that can take three values: 1 for \textit{active}, -1 for \textit{inactive}, and 0 which can be interpreted as \textit{unobserved, undefined,} or \textit{irrelevant} depending on context. Note that the numerical values are given only as shorthand notation and do not participate in an algebraic operation anywhere. The phrase \textit{nonactive} refers to any SV that is not active. The SV construct comes in three subtypes: Base SVs (BSVs), Dynamics SVs (DSVs), Conditioning SVs (CSVs).

\textit{BSV:} BSVs are the externally-specified SVs whose states, which is assumed to be either 1 or -1, are provided externally to the system at each time instant. These can be regarded as the direct observations from the environment.

\textit{DSV:} Each BSV comes with two associated DSVs, for activation (A-DSV) and deactivation (D-DSV) respectively. Activation at timestep $t$ is defined as the transition of a BSV state from -1 in step $t-1$ to 1 in step $t$; and likewise deactivation at $t$ is defined from 1 in $t-1$ to -1 in $t$. At step $t$, A-DSV is deduced active (state 1) if activation is observed at step $t$, inactive (-1) if a BSV is inactive at $t-1$ and no activation is observed at $t$, and undefined (0) if the BSV is already active. Symmetrically, at step $t$, D-DSV is deduced active (state 1) if deactivation is observed at step $t$, inactive (-1) if a BSV is active at $t-1$ and no deactivation is observed at $t$, and undefined (0) if the BSV is already inactive. The BSVs are modelled only through changes in their states via their associated DSVs, and are not predicted by themselves.

\textit{CSV:} A CSV is a SV that conditions either DSVs or other CSVs (but not BSVs since they are not subject to direct modelling of their states); that is, predicts their activation. More specifically; each CSV comes with a set of positive and negative sources, where each source is either a BSV or DSV; and a set of targets, which correspond to the SVs that this CSV conditions. At steady state, a CSV’s source conditions are said to be satisfied when all its positive sources were active and all its negative sources were nonactive in the previous step - in other words, the satisfaction corresponds to the condition $all(positive\ sources)\ and\ not(any(negative\ source))$ in the previous step. A CSV state is undefined (0) if its source conditions are not satisfied. If its source conditions are satisfied; a CSV’s state is active (1) if the state of all its targets are either active or unobserved; and inactive (-1) if the state of all its targets are either inactive or unobserved. In case inactive and active targets are observed together, the CSV is duplicated to encompass the corresponding subsets of targets (as detailed below), hence we always ensure that one of the two above conditions will be satisfied with respect to the states of the targets. A CSV is to be interpreted as a state variable that represents the observance of a particular relationship - it being active means that this particular relationship (e.g. a change, as represented by a DSV, is observed conditioned on some sources) is observed, and it being inactive means that this relationship is not observed. The CSV being undefined or unobserved corresponds to the case in which the conditions for the observation of the relationship are not satisfied in the first place.

Potential targets of conditioning (i.e. DSVs and CSVs), when they are not undefined, are expected to be active if one of their conditioners are active; and inactive otherwise. Furthermore, these types of SVs also possess an \textit{unconditionally} flag, that allow for exceptions in this activity prediction, and are used to model uncertainty regarding activation of SVs. This flag can take three values: It starts with a value "unconditional" at the creation of the CSV and, if the CSV is observed to always be active whenever its sources were satisfied, it remains so. At the first observation of a case where the sources of the CSV are satisfied without the CSV being active, this flag changes to "conditional," signalling that sources alone do not suffice for the activation of the CSV and activity of one of its upstream conditioners is expected. The "conditional" value persists until the first observation of a case where CSV is observed active without any upstream conditioner being active and no new conditioner could be formed (see below and the main text); in which case the flag changes to "possibly unconditional" and remains as such.

Over the course of interaction with the environment, Modelleyen learns a model that predicts the BSV states at the next step indirectly via the prediction of the DSV states. Within the predictions uncertainty is also represented where needed, as apparent from the description of the SVs. Since uncertainty is represented in a local basis (by unconditionality flags of individual SVs), and since CSVs are points of connection relating potentially multiple sources to potentially multiple targets; the uncertainty representation can represent alternative correlated outcomes in a tree-like manner where each downstream “branch” corresponding to the alternative outcomes in one direction or another can include multiple outcomes that occur together - we note that representation of uncertainty as such is not possible in a local manner with e.g. classical neural networks.

\subsection{Learning the model}
\label{sec:modelleyen_details_learning}

First, we provide an overview of the learning process in one step of interaction with the environment. During a step, the model is traversed, and the states of all its SVs are computed. For CSVs sources and targets are modified to be able to match the current states to the predictions/explanations of the CSV, so that the model is consistent with the environment at each step. After that, new CSVs are generated for the DSVs and CSVs that lack an explanation at the current step. The new CSV takes as positive sources all currently active eligible SVs in an exhaustive manner. Finally, model is refined by removal of unnecessary state variables.

The learning process is summarized formally on Algorithms \ref{alg:algorithm_adaptationloop} and \ref{alg:algorithm_csvstate}. Below, we provide a detailed breakdown of the processes described on those algorithms.

\begin{algorithm}[tb]
\caption{Pseudocode of the main Modelleyen adaptation loop; formed of state computations followed by CSV generation for unexplained SVs.}

\label{alg:algorithm_adaptationloop}
\textbf{Parameter}: $N$ Set of all target nodes\\
\textbf{Function} \textit{ProcessEnvironmentStep}(observations)
\begin{algorithmic}[1] 
    \STATE $BSVStates \leftarrow \ observations$
    \STATE $ComputeDSVStates()$ \COMMENT{Computes DSV states by BSV events}
    \FOR{$level\ \in\ reverse(ComputationLevels)$}
        \FOR{$CSV\ \in\ SVs_in(level)$}
            \STATE $ComputeState(CSV)$
        \ENDFOR
    \ENDFOR
    \STATE $UnexplainedSVs \leftarrow [SV:\ SV.state = 1\ \AND\ NoConditionerActive(SV)]$
    \STATE $sources \leftarrow [SV:\ SV in\ [BSVs, DSVs]\ \AND\ SV.state = 1\ \AND\ isEligible(SV)]$
    \STATE $ NewCSV = CreateCSV(sources, [SV: SV\ in\ UnexplainedSVs\ \AND\ TargetEligible(SV)])$
    \STATE $ModelRefinement()$ \COMMENT{Removes CSVs with no source or target}
\end{algorithmic}
\end{algorithm}

\begin{algorithm}[tb]
\caption{Pseudocode for CSV state computation.}

\label{alg:algorithm_csvstate}

\textbf{Function} \textit{ComputeState}($CSV$)
\begin{algorithmic}[1] 
    \IF{$AnySourceActive()$}
        \STATE $SeparateActiveInactiveTargets()$ \COMMENT{Creates two CSVs from current one with active and inactive targets in either of them}
        \IF{$AnyTargetObserved()$}
            \STATE $State = 1$
            \STATE $PosSources \leftarrow [source:\ source\ in\ PosSources\ \AND\ source.state=1]$
            \STATE $NegSources \leftarrow [source:\ source\ in\ NegSources\ \AND\ source.state!=1]$
        \ELSIF{$AnyTargetInactive()$}
            \IF{$not(AllSourcesActive())$}
                \STATE $State = 1$
            \ELSE
                \IF{$AnyNegativeSourceActive()$}
                    \STATE $State = 0$
                    \STATE $NegSources \leftarrow [source:\ source\ in\ NegSources\ \AND\ source.State=1]$
                \ELSE
                    \STATE $State = -1$  \COMMENT{No negative source active to explain inactivity of targets}
                \ENDIF
            \ENDIF
        \ENDIF
    \ELSE
        \STATE $State = 0$ \COMMENT{Unobserved if targets are not observed}
    \ENDIF

    \IF{$State = -1$}
        \IF{$NegativeConnectionsFormed$}
            \STATE $FormNegativeConnections()$
        \ELSE
            \STATE $unconditionality = "isConditional"$ \COMMENT{-1 for }
        \ENDIF
    \ENDIF

\end{algorithmic}

\end{algorithm}

Initially, the model is generated with only BSVs and their associated DSVs, and without any CSV. At every step, the current and previous states of all the SVs are recorded, as well as the current and previous events (activation and deactivation) of every BSV. 

At each step, the effective network created by DSVs and CSVs are traversed in the reverse order of computation, similar to backpropagation algorithm; starting from DSVs, then the CSVs that condition these BSVs, then the conditioners of these CSVs, and so on. Each traversed SV gets their state computed, and additionally CSV compositions are changed where needed, as in Figure \ref{fig:csvform} and detailed below.

\subsubsection{Processing of a CSV}

The process for CSVs are carried as follows: If no positive source of a CSV is observed at a given step, its state is deduced as 0 (undefined/unobserved). If at least one source is observed, and if there are both active and inactive targets among the CSV targets, then the CSV is duplicated with different target sets to create one copy that includes active targets and one copy that includes inactive targets (and any undefined targets are shared by both). This ensures that the CSV remains consistent, since it’s activation represents the activation of all its targets provided they are not undefined. There is no way to say whether an undefined target will be consistent with one duplicate or another after the changes to the CSV described below without observing a non-undefined state in them, so they are put into both copies and do not otherwise affect the state deduction of the CSV (except if all targets are undefined, see below).

Following this operation, if a CSV has any target active, then its state is deduced as active (1). If there is no perfect match with the standing sources of CSV and their activations (i.e. there are either inactive positive sources or active negative sources), these source lists are refined so that the remaining sources correspond perfectly to the current state of the network - in other words, any positive source that is inactive and any negative source that is active is removed. This refinement eliminates parts of the previously-posited relationships “hypothesized” to be necessary by the CSV in an exhaustive manner (see details on CSV formation, below) that are observed to be not necessary for the observation of the effect that the CSV models (Figure \ref{fig:csvform_3}.

If, on the other hand, the CSV has any inactive target (which is exclusive with any target being active due to the duplication-differentiation operation made above) and if not all its positive sources are active, then the state is deduced as 0, being consistent with the interpretation of a CSV as being defined only if all its positive sources are active. If however, all positive sources are active; then we look if any negative source is active that can justify the inactivation of the targets of the CSV. If there is at least one negative source that is active, we deduce the state as 0 since source conditions are not satisfied; and refine the negative targets that are not currently active in the same manner we described in the previous paragraph (due to the observation that they are seen to be not necessary for the suppression of the CSV - Figure \ref{fig:csvform_5}).

If, instead, all the targets of CSV are undefined, then the CSV is undefined as well.

A CSV is always created with only positive sources at first and no negative sources, and a CSV always starts as an unconditional CSV for whom we never expect to observe an inactive state (see below part for details on the generation of CSVs). At the observation of an inactive state in the CSV (i.e. one in which sources are active but targets are inactive), only once after the creation of the CSV, we duplicate the CSV and separate the targets that are currently undefined (to protect them from the change being made). In the duplicate that has the inactive targets, we connect the CSV with the negative sources by forming a negative sources list that encompasses all the currently-active eligible BSVs and DSVs in the model, which will be subject to future refinement (criteria of \textit{eligibility} is detailed in the Appendix, essentially corresponding to SVs that do not yield useful information). This, essentially, attempts to explain the CSV’s observed inactivation. If, however, an inactive state is observed despite already having formed connection with negative sources, then the unconditionally flag of the CSV is set to "conditional", representing that the CSV’s state is now uncertain (setting aside its possible conditioners).

\subsubsection{CSV generation and model refinement}
\label{subsec:csv_form}

After the traversal of SVs for computation of their states and modifications in CSV compositions, all DSVs and CSVs who are observed active but are neither unconditional nor have an active conditioner that explains their activation are labelled as \textit{unexplained}. We then form a CSV that, as positive sources, has all the eligible, currently-active BSVs and DSVs; and as target, has all the eligible SVs in unexplained list (Figure \ref{fig:csvform_1}). Any target which is left outside of this CSV, and hence remain unexplained, have their unconditionally flags set to "possibly conditional" (which basically signals that the SV can go active without any explanation or predictor).

Finally, at the end of the step, we refine the general model by removing any CSVs that may be duplicates of other CSVs (ending up representing the same thing from different histories), as well as any CSV that has no sources or targets left as a result of refinement or duplication operations.

\subsubsection{Source eligibility for CSVs}

To reduce model complexity and avoid the need for repeated exposures to the environment, we pre-filter sources during CSV formation or CSV negative-sources formation by their eligibility as follows: We define \textit{trivial sources} of a CSV as the sources of all the SVs that lie downstream starting from this CSV (i.e. SVs conditioned by this CSV, and CSVs conditioned by them, and so on), plus the associated BSV if a DSV is reached. Intuitively, these are the sources whose states can be determined by the knowledge that the CSV is active (since a CSV being active means that it’s target will be active as well, which will inform us about the states of its sources), and hence wouldn’t be informative sources for the current CSV as any information conveyed by them will be trivial. When forming a CSV, among all the currently-active BSV and DSVs, we filter those that provide trivial information to all the unexplained SVs (i.e. prospective targets for the generated CSV) out as positive sources, and take only those that do not provide trivial information as source to at least one of them. Furthermore, after this filtering, if there is a prospective target for which all the remaining prospective sources provide trivial information, then this target is not taken as a target of the CSV and hence remains unexplained.

In a similar spirit, when forming negative sources, we filter out all the candidates that provide trivial information for the CSVs. In addition, however, we filter out any upstream positive source (that is, the cumulative list of all positive sources among all upstream CSVs of this CSV, i.e. its conditioners and conditioners of its conditioners, including itself) because we already know (by the definition of the conditioning process) that there was an instance in which this CSV was observed when the SVs in this list of positive conditioners was also observed; and hence these negative sources would be eliminated in exposure with the same instance again.

\subsubsection{Conditioner formation for unconditional CSVs}

Here we note a modification that we do not employ currently, but is possible: Currently we allow no CSVs to condition unconditional CSVs since they are not informative and hence prevent the model from being minimal. However, we note that allowing for conditioners to be formed to unexplained (no active conditioners)  unconditional CSVs as well could result in these CSVs already having some conditioners learned from the previous encounters with the environment in case they ever turn conditional, reducing the required number of interactions for the learning of the full environment model, at the cost of making the model more exhaustive in terms of what is being modelled. This would require two changes: (1) At CSV formation, not excluding the unexplained CSVs that are unconditional; and (2) when refining positive sources, we create a CSV which takes as its initial positive sources that are being removed, and that conditions the CSV whose sources are being refined currently. This way, instead of removing what was observed to be active at previous encounters at which the CSV was active, we push them to an upper level of computation to represent an alternative condition in which the CSV was observed to be active before.

\subsection{Proof of Theorem 1}
\label{sec:app_proof}

Let $X_P^i$ and $X_N^i$ be positive and negative sources of $C$ respectively that remains \textit{after} refinements that instance $y_i$ causes. Since we know that $C$ does not undergo negative sources formation, and that $y_0$ comes before $y_1$, we can say that $X_P^1 \subseteq X_P^0$ and $X_N^1 \subseteq X_N^0$ since only refinements are allowed on $X_P$ and $X_N$ sets of $C$ by our definition of operations.

We now analyse the two possible cases with respect to satisfaction of sources:
\begin{itemize}
    \item If, in the original encounter with $y_0$ the sources of $C$ were satisfied, then we had $S_x=1 \forall x \in X_P^0$ and $S_x=1 \forall x \in X_P^0$. Since $X_P^1 \subseteq X_P^0$ and $X_N^1 \subseteq X_N^0$, we will also have $S_x=1\ \forall x \in X_P^1$ and $S_x=1\ \forall x \in X_P^1$ at the new encounter with instance $y_0$. Hence, if sources of $C$ were satisfied in the previous encounter with $y_0$, they will remain satisfied in the new encounter. The value of $S_C$ can be -1 or 1 if and only if sources of $C$ are satisfied; in which case it is exclusively determined by the state of its targets (-1 if targets are inactive and 1 if targets are active). Since the states of targets are determined by $y_0$ and hence is the same across the past and new encounter with $y_0$; if $S_C=1 (-1)$ in the past exposure with $y_0$, then it will be $1(-1)$ in the new exposure as well.
    \item If, in the original encounter with $y_0$ the sources of $C$ were not satisfied (and hence original encounter yielded $S_C=0$), then we either had $S_x \neq 1\ \forall x \in X_P^0$ or $S_x = 1\ \forall x \in X_N^0$ (note that we defined $X_P^i$ and $X_N^i$ as source sets \textit{after} the refinements; and hence we know that in both cases it will be the whole of positive/negative source sets that have the property, and not a subset of them; since the source SVs that were not a part of that subset will have been refined). Since $X_P^1 \subseteq X_P^0$ and $X_N^1 \subseteq X_N^0$, we will also have either $S_x \neq 1\ \forall x \in X_P^1$ (if former) or $S_x = 1\ \forall x \in X_N^1$ (if latter), both of them not satisfying the sources conditions of $C$ (hence the new encounter with $y_0$ also yielding $S_C=0$.
\end{itemize}

Therefore, in all cases, response to $y_0$ remains identical before and after exposure to $y_1$.

\subsection{Learning the statistical significance of encountered relations}
\label{sec:statistical_significance}

The base mechanisms of Modelleyen as described in the main text rest on an attempt of prediction of all encountered changes in state variables in the environment, forming an explanatory/predictive relationship between any two observed events in that attempt of full modelling of the environment. Unlike neural networks (or other statistical learning methods), the naive algorithm does not depend on, but also does not naturally incorporate, a method of statistically averaging and filtering learned relationships. Such a means of estimation of statistical significance of learned relationships can be incorporated into the models learned by modelleyen in a straightforward manner into the learned relationships locally, which in turn can be used to filter out non-significant relationships, hence preventing overcomplexification of the model.

Let $C$ be a CSV, and let $T$ be a target SV of that CSV. We define the event \textit{sources satisfied}, $SS(C)$, to be the event where all positive sources of $C$ are active and all negative sources are nonactive. For each target, we define an \textit{observation} of the target $O(T)$ to be when the target is observed (i.e. either active or inactive, state 1 or -1, as defined in the main text) and an \textit{incidence} of the target $I(T)$ to be when the target is active (state 1). We define the event \textit{concurrence} to be the event where both the sources of $C$ are satisfied and there is an indicence of target, $CC(C,T)=SS(C) \wedge I(T)$.

We quantify the statistical significance of a learned relationship between a set of sources of a CSV and one of its targets as the \textit{amount of increase in the probability of the incidence of the target given the satisfaction of the sources of the CSV}. We define \textit{normalized causal effect (NCE)} as the amount of increase in probability of incidence of $T$ that satisfaction of sources of CSV $C$ causes, normalized by the original probability of incidence:

\begin{equation}
    NCE = \frac{P(I(T)|SS(C)) - P(I(T))}{P(I(T))}
\end{equation}

The conditional probability in the nominator can be expanded as:

\begin{equation}
    P(I(T)|SS(C)) = \frac{P(I(T), SS(C))}{P(SS(C))} = \frac{P(CC(C,T))}{P(SS(C))}
\end{equation}

by our definition of concurrence $CC(C,T)$ above. All of the probabilities can be computed by locally tracking of the number of instances that the corresponding events are observed, when the target is observed (i.e. $O(T)=1$). When the target is unobserved/undefined, by extension none of the other events are observed.

A positive NCE means that $SS(C)$ increases probability of $I(T)$ and a negative NCE means that $SS(C)$ decreases it. An NCE of e.g. 2.0 means that $SS(C)$ increases probability of $I(T)$ to 3 times the original probability. Within the context of our modelling mechanism, a negative NCE means that the relationship between sources of $C$ and $T$ has been learned in the wrong direction - actual negative relations learned in proper direction will still result in positive NCE, because the sources of that relation will go within the negative sources of $C$ instead of the positive ones, still in the end resulting in the $SS(C)$. The lower the magnitude of NCE, the less significant the relationship is.

Given NCE values for each relationship, one can set a positive threshold $\epsilon_T$, where NCE values with magnitude below it are regarded as statistically insignificant. $\epsilon_T$ represents the trade-off between complete modelling and model complexity. After that separation of relationships into significant and insignificant ones, one can proceed either with their removal, or simply with blocking further conditioner formation for them to prevent overcomplexification in an attempt to predict a near-random relationship (i.e. to prevent "fitting the noise"). Since our main aim in employing this mechanism is to prevent overcomplexification, and since removal of such insignificant relationships from the model completely would result in their re-learning if the agent is exposed to them again; we opt for the latter option and block further conditioner formation for them.

NCE values may have other utilities for the processes of the agent. An example might be that it can be used in the prioritization of subgoals in Planlayan (see main text), where more reliable causal relationhips are prioritized over less reliable ones. We do not investigate into such utilities at this stage.

It’s important to note that the statistical estimates are not precise during the transient phase. This is due to the refinement mechanism, which prioritizes structural revisions and adjustments to make a given CSV align with observations where feasible. During this phase, estimates tend to overemphasize significance. However, these transients are brief, and NCEs insignificant CSVs quickly diminish once the refinements are complete and the CSV sources settle into their final form. Furthermore, this final form is typically less constrained, leading to more exposures over time in the same environment. Alternatively, we could eliminate these inaccuracies by resetting recorded statistics after each change to the CSV's composition, though this would increase the time needed for an NCE value to be deemed reliable. We do not use this approach here, as we do not find the temporary bias toward significance in transient SVs to be an issue, but it can be employed where precision has priority over efficiency.

\paragraph{Effect on continual learning} Notice that there is no change (particularly no decay) in NCE if the target is not observed - hence, this measure of statistical significance does not decay (relationship "forgotten") in case of a changed environment in which the new one does not display the co-occurance of the two events (target and CSV sources being satisfied), as long as its target is not observed in isolation as well. If its target is observed in the new environment, two cases may occur:
\begin{enumerate}
    \item $P(I(T))$ is stable. This would be expected in an already-mature model or in environments where there is not much variability in the occurance of individual targets (even if the conditions under which they occur differ). In this case, there is no change in NCE.
    \item $P(I(T))$ changes. In this case, NCE will change according to $P(I(T))$. Note, however, that additional exposure can only mean a more accurate estimate of the true $P(I(T))$ value - any change in $P(I(T))$ hence does not have a detrimental effect, but instead makes the causal effect estimate more reliable in the context of the complete model; provided that the new environment itself does not have a probability of $P(I(T))$ in itself that is non-representative of the general probability, in particularly one that is excessively higher than the general one. This latter possibility (an immature estimate of $P(I(T))$ and an unnaturally high $P(I(T))$ in the new environment) is the only case in which a previously-learned correct relationship can be wrongly destroyed in case of a changing environment. But even such cases would have no long-term ramifications as $P(I(T))$ for any given target $T$ would reach to a reliable estimate after a few cycles of exposures to environments where $T$ is observed.
\end{enumerate}

The current method of computing and filtering based on statistical significance has one drawback, however; and it is that only first-order significance of relations are considered. In other words: If we have a CSV C0 with a target D0, and C0 (possibly unconditional) is conditioned by another CSV C1, then whether C0-D0 relationship will be regarded as significant or not depends only on the observations of sources of C0 and D0; and will \textit{not} consider their dependency on C1. This may result in unnecessary filtering in cases where a said statistical relationship is insignificant in the absence of a particular upstream conditioner, but becomes significant with that - we also see effects of this limitation to some degree in our results in the main text. Resolution of this limitation requires consideration of and conditioning on higher-order conditioners when computing the NCE value, and is left for future work.

\subsection{Details of experimental framework}
\label{sec:appendix_expsetup}

\paragraph{Significance filtering} Modelleyen's mechanism of filtering based on statistical significance (i.e. NCE) is enabled only for the random variant of the environment. When enabled, we used a cutoff NCE of 0.25 for blocking upstream conditioner formations (i.e. no more upstream conditioners are formed if the CSV does not cause a >25\% in the probability of occurrence of its target).

\paragraph{Intuition regarding the design of environment in Figure \ref{fig:environment}} The environment was inspired from Multiroom environment in Minigrid. The states represent closed door (DC), open door (DO), wall (W), subgoal 1/2 (SG1/2), goal (G) and a random variable (X); "RS" stands for "rooms" and represents an agent going through multiple rooms opening doors in each, and "SGS" represents one in which agent reaches two subgoals and then reaches the goal afterwards, and "NEG" represents a case where goal appears conditioned on one positive and one negative conditon. In all, the goal can be moving. Alternative outcomes are present in all environment subtypes, since each of them allows for multiple outcomes following an empty ("-/-") state. Alternative predecessors are tested in "SGS" environment where SG2 can be preceded by SG1 in either of the two cells; and likewise in general the appearance of G can be preceded by any of the alternatives associated with different environment subtypes. The capability to represent positive and negative relations together is tested in subtype "NEG", in which G appears only if X is enabled in the first cell and not the second one.

\paragraph{Computation resources} All experiments were run on a 2.4GHz 8-Core Intel Core i9 processor with 32 GB 2667MHz DDR4 memory. No GPU was used. Giving an accurate estimate for computation time is impossible since experiments were run in parallel to unevenly-distributed independent workloads.

\subsection{A sample model learned on SMR}

\begin{figure*}
  \centering
  \includegraphics[width=0.5\textwidth]{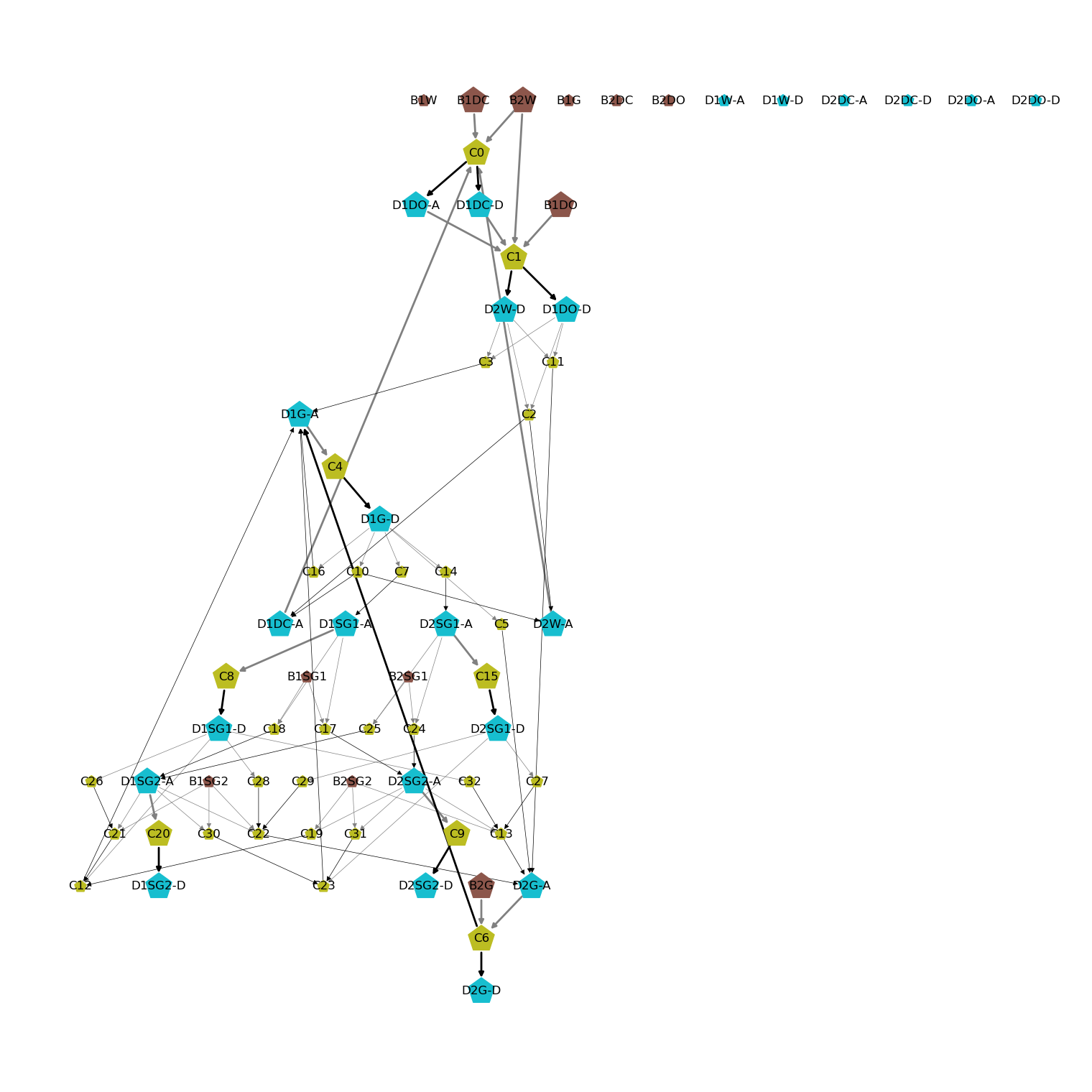}
  \caption{A sample environment model learned by Modelleyen. In the visualized model, brown nodes are BSVs, blues are DSVs, and the rest are CSVs. The enlarged pathways (bold arrows and large nodes) are reliable outcomes (i.e. unconditional CSVs) and the rest are uncertain (possibly conditional) ones. Black arrows represent conditioning relationships and gray arrows represent source relationships (all positive in this example). Disconnected SVs (those that can never be activated by environment design) are cut for visual clarity.}
  \label{fig:sample_model}
\end{figure*}

\begin{figure}
  \centering
  \includegraphics[width=0.5\textwidth]{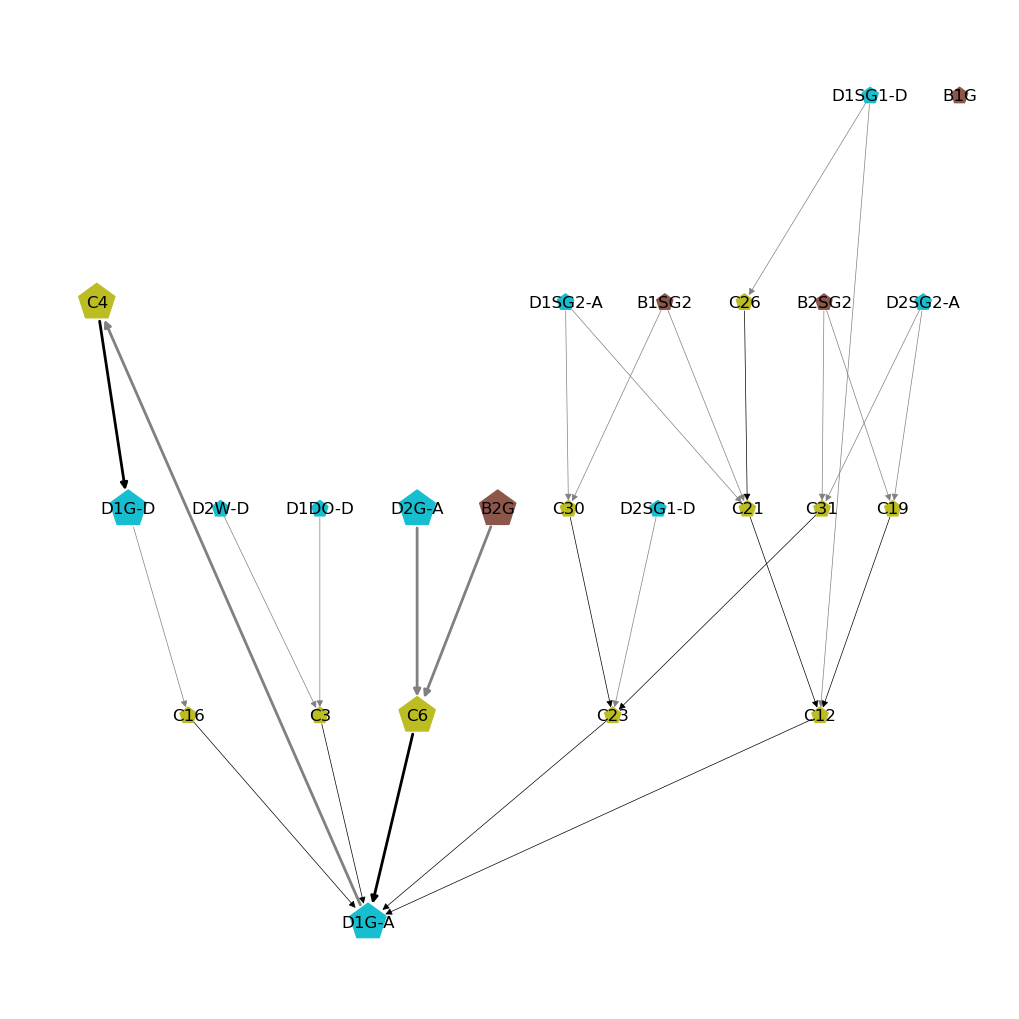}
  \caption{Same model as Figure \ref{fig:sample_model}, but for the predictive pathway of BSV 1G only. Many pathways for the activation of 1G can be seen in a human-comprehensible way in this model via the distinct CSVs preceding it (C3, C6, C12 C16, C23) and that the only reliable one of them is C6, and whose further sources can be seen by pursuing them upstream. In contrast, interpretation of a neural network model is much less straightforward due to nonlinearities, continuous parameters, and extensive connectivity that ties each neuron at the output to virtually all other neurons in the network.}
  \label{fig:sample_model_goal}
\end{figure}

\begin{figure*}
  \centering
  \includegraphics[width=0.5\textwidth]{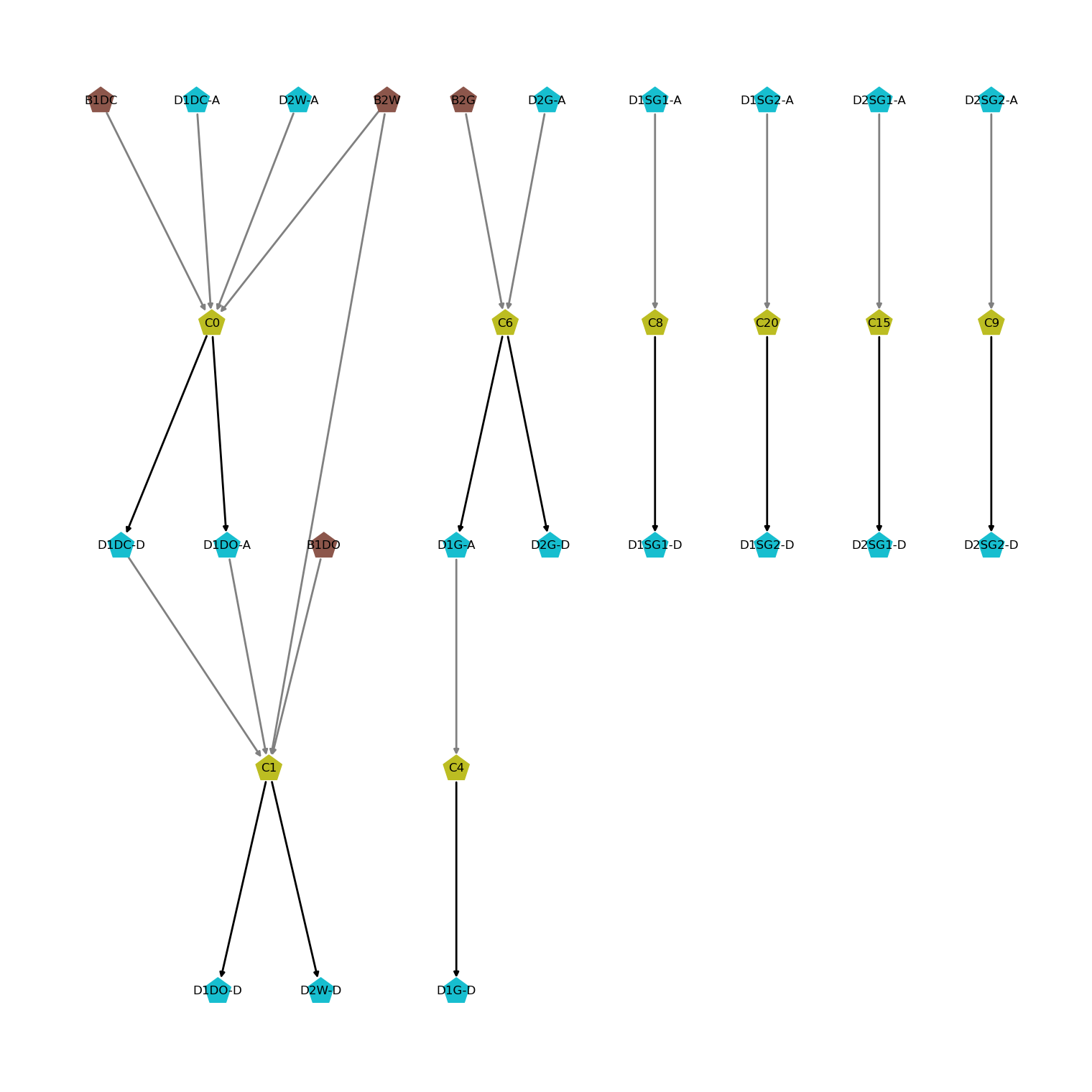}
  \caption{Same model as Figure \ref{fig:sample_model}, but with reliable pathways only, showing "islands of certainty" as potential candidates for abstraction.}
  \label{fig:sample_model_reliable}
\end{figure*}

A sample model learned on the SMR environment (Figure \ref{fig:environment}) is provided on Figure \ref{fig:sample_model}. Figure \ref{fig:sample_model_goal} provides, as an example, the pathway of BSV 1G (state G at cell 1), in which the specific pathways connecting to this BSV can be seen more clearly in a human-comprehensible manner. Figure \ref{fig:sample_model_reliable} shows the whole model, but only with reliable connections; clearly showing "islands of certain state transitions" which can be an example of a delimiting criterion that can be used for abstractions as discussed in the main text.

\begin{algorithm}[tb]
\caption{Simplified overview of the planning algorithm, relying on recursive generation of upstream \textit{action networks} (the graph of behaviors required to realize the desired goals from the currently active SVs).}

\label{alg:planner}
\textbf{Function} Plan(currentActiveSVs, goalSVs)
\begin{algorithmic}[1] 
    \STATE ActionNetwork $\leftarrow$ \ EmptyNet
    \FOR{SV, target $\in$ goalSVs}
        \STATE GenerateUpstreamAN(SV, target)
    \ENDFOR
\end{algorithmic}

\textit{Comment: Argument "target" states what the desired state is in the SV, which can be activation (A), deactivation (D), active (1) or nonactive (0). Irrelevant for CSVs.}

\textbf{Function} GenerateUpstreamAN(SV, target)
\begin{algorithmic}[1] 
    \STATE \textbf{if} satisfiedByCurrentActives(SV, target): return \textbf{True} 
    \STATE pathways $\leftarrow$ EmptyList
    \IF{type(SV) in [BSV, GSV]}
        \STATE pathways.add(Precondition(sv, target))
        \STATE{\textit{Comment: These are the preconditions for target to occur in a SV. For (A, D, 1, 0) they are (0, 1, A, D) respectively; since a SV must be activated for itself to be active, needs to be inactive for itself to get activated, and so on.}}
        \STATE pathways.add(Constituents(sv), target)
        \STATE pathways.add(Constituencies(sv), target)
        \STATE \textbf{if} target in ['A', 'D']: pathways.add(Conditioners(sv, target))
    \ELSIF{type(SV) is CSV}
        \STATE pathways.add(Sources(sv))
        \STATE pathways.add(Conditioners(sv))
    \ENDIF
    \STATE \textbf{if} pathways is Empty: return \textbf{False}

    \FOR{upstreamSV, upstreamTarget in pathways}
        \STATE ActionNetwork.AddEdge((upstreamSV, upstreamTarget), (SV, target))
        \STATE GenerateUpstreamAN(upstreamSV, upstreamTarget)
    \ENDFOR
\end{algorithmic}
\end{algorithm}

\end{document}